%% file: main.tex
\documentclass[a4paper,twoside]{article}

\input{Setup/Statics.tex}
\input{Setup/Preamble.tex}
\input{Figures/tikzstyle}

\usepackage{SCITEPRESS}

\begin{document}

\title{Membership Inference Attacks for Face Images Against Fine-Tuned Latent Diffusion Models}

 \author{\authorname{Lauritz Christian Holme\sup{1}\orcidlink{0009-0001-3043-5561}, Anton Mosquera Storgaard \sup{1}\orcidlink{0009-0001-1437-6004} and Siavash Arjomand Bigdeli\sup{2}\orcidlink{0000-0003-2569-6473}}
\affiliation{\sup{1} Department of Applied Mathematics and Computer Science
, Technical University of Denmark, Kongens Lyngby, Denmark}}


\abstract{The rise of generative image models leads to privacy concerns when it comes to the huge datasets used to train such models. This paper investigates the possibility of inferring if a set of face images was used for fine-tuning a Latent Diffusion Model (LDM). A Membership Inference Attack (MIA) method is presented for this task.  Using generated auxiliary data for the training of the attack model leads to significantly better performance, and so does the use of watermarks. The guidance scale used for inference was found to have a significant influence. If a LDM is fine-tuned for long enough, the text prompt used for inference has no significant influence.  The proposed MIA is found to be viable in a realistic black-box setup against LDMs fine-tuned on face-images.}

\onecolumn \maketitle \normalsize \setcounter{footnote}{0} \vfill

\section{\uppercase{Introduction}} \label{sec:introduction}
\input{Sections/1_introduction}

\section{\uppercase{Related Works}} \label{sec:related_works}
\input{Sections/2_related_works}

\section{\uppercase{Data}} \label{sec:Data}
\input{Sections/3_data}

\section{\uppercase{Method}} \label{sec:Method}
\input{Sections/4_method}

\section{\uppercase{Results}} \label{sec:Results}
\input{Sections/5_results}

\section{\uppercase{Conclusion}} \label{sec:conclusion}
\input{Sections/6_conclusion}

\bibliographystyle{apalike}
{\small
\bibliography{bibliography}}

\end{document}

%% file: Setup/Statics.tex
\newcommand{\hwmseen}{$\mathbf{\leftindex^{\mathbf{hwm}} {D}_{DTU}^{seen}}$}
\newcommand{\hwmgen}{$\mathbf{\leftindex^{\mathbf{hwm}} {D}_{DTU}^{gen}}$ }
\newcommand{\wmseen}{$\mathbf{\leftindex^{\mathbf{wm}} {D}_{DTU}^{seen}}$ }
\newcommand{\wmgen}{$\mathbf{\leftindex^{\mathbf{wm}} {D}_{DTU}^{gen}}$ }
\newcommand{\hwmtar}{$\mathbf{\leftindex^{\mathbf{hwm}} {M}_{T}^{DTU}}$}
\newcommand{\wmtar}{$\mathbf{\leftindex^{\mathbf{wm}} {M}_{T}^{DTU}}$}

\DeclareMathAlphabet{\mymathbb}{U}{BOONDOX-ds}{m}{n}

%% file: Setup/Preamble.tex
\usepackage{epsfig}
\usepackage{subcaption}
\usepackage{calc}
\usepackage{amssymb}
\usepackage{amstext}
\usepackage{amsmath}
\usepackage{amsthm}
\usepackage{multicol}
\usepackage{pslatex}
\usepackage{apalike}
\usepackage{algorithm2e}
\usepackage[bottom]{footmisc}

\usepackage{algpseudocode}
\usepackage{multirow}
\usepackage{graphicx}
\usepackage{icomma}
\usepackage{multirow}
\usepackage{pgf}
\usepackage{pgffor}
\usepackage{pgfplots}
\usepgfplotslibrary{groupplots,dateplot}

\usepackage{changepage}
\usepackage{threeparttable}
\usepackage[]{geometry}     
\usepackage{tikz}           
\usetikzlibrary{shapes.geometric, arrows.meta, shadows, positioning, patterns}
\colorlet{myred}{red!20!white}
\colorlet{myblue}{blue!20!white}
\colorlet{mygreen}{green!40!white}
\colorlet{mypurple}{blue!50!red!50!white}
\colorlet{mydarkred}{myred!40!black}
\colorlet{mydarkblue}{myblue!40!black}
\colorlet{mydarkgreen}{mygreen!40!black}
\tikzset{no shadows/.style={general shadow/.style=}}

\pgfplotsset{compat=newest}
\usepackage{etoolbox}
\usepackage{listofitems}
\usepackage{pgfplots}       
\usepackage{xcolor}         
\PassOptionsToPackage{hyphens}{url} 
\usepackage{hyperref}       
\usepackage{orcidlink}
\usepackage{cleveref}       
\usepackage{textcomp}       
\usepackage[english]{babel} 
\usepackage{caption}        
\usepackage{csquotes}       
\usepackage{tabularx}       
\usepackage{booktabs}       
\usepackage{float}          
\usepackage{calc}           

\usepackage{leftindex}      

\usepackage{pdflscape}
\usepackage{pdfpages}

\usepackage[toc,section=section, acronym]{glossaries} 
\input{Sections/Glossary}

%% file: Sections/Glossary.tex
\makeglossaries

\newacronym{DTU}{DTU}{Technical University of Denmark}
\newacronym{BERT}{BERT}{Bidirectional Encoder Representations from Transformers}
\newacronym{GPT}{GPT}{Generative pre-trained transformer}
\newacronym{API}{API}{Application programming interface}
\newacronym{TA}{TA}{Teaching Assistant}
\newacronym{AI}{AI}{Artificial Intelligence}
\newacronym{NLP}{NLP}{Natural Language Processing}
\newacronym{LM}{LM}{Language Model}
\newacronym{LLM}{LLM}{Large Language Model}
\newacronym{ANN}{ANN}{Artificial Neural Network}
\newacronym{ML}{ML}{Machine Learning}
\newacronym{ROC}{ROC}{Receiver operating characteristic}
\newacronym{AUC}{AUC}{Area Under the Curve}
\newacronym{MIA}{MIA}{Membership Inference Attack}
\newacronym{LDM}{LDM}{Latent Diffusion Model}
\newacronym{GDPR}{GDPR}{General Data Protection Regulation}
\newacronym{AAU}{AAU}{Aalborg University}
\newacronym{VAE}{VAE}{Variational Autoencoder}
\newacronym{CLIP}{CLIP}{Contrastive Language–Image Pre-training}
\newacronym{PCA}{PCA}{Principal Component Analysis}
\newacronym{LFW}{LFW}{Labeled Faces in the Wild}
\newacronym{FID}{FID}{Fréchet Inception Distance}
\newacronym{GAN}{GAN}{Generative Adversarial Network}

%% file: Figures/tikzstyle.tex
\tikzstyle{arrow} = [thick,->,>=stealth]
\tikzstyle{arrow_t} = [ultra thick,->,>=stealth]
\tikzstyle{arrow2} = [thick,-,>=stealth]
\tikzstyle{arrow3} = [thick, ->, >=stealth, dashed]

\tikzstyle{normal} = [rectangle, align=center, rounded corners, minimum width=1cm, minimum height=1.0cm, text centered, draw=black, fill=white, drop shadow]

\tikzstyle{normal_un_cen} = [normal, align=left]

\tikzstyle{black} = [rectangle, align=center,  rounded corners, minimum width=1cm, minimum height=1.0cm, text centered, draw=black, fill=black!80, text=white, drop shadow]

\tikzstyle{attack} = [rectangle, align=center,  rounded corners, minimum width=1cm, minimum height=1.0cm, text centered, draw=black, fill=mypurple, text=black, drop shadow]

\tikzstyle{positive} = [rectangle, align=center,  rounded corners, minimum width=1cm, minimum height=1.0cm, text centered, draw=black, fill=myblue, text=black, drop shadow]

\tikzstyle{negative} = [rectangle, align=center,  rounded corners, minimum width=1cm, minimum height=1.0cm, text centered, draw=black, fill=myred, text=black, drop shadow]

\tikzstyle{image} = [rectangle, align=center, rounded corners, minimum width=1cm, minimum height=1.0cm, text centered, draw=black, fill=white, drop shadow]

\tikzstyle{detail} = [rectangle, dashed, align=center, rounded corners, minimum width=1cm, minimum height=1.0cm, text centered, draw=black, fill=white]

\tikzstyle{special} = [diamond, align=center, rounded corners, minimum width=1cm, minimum height=1.0cm, text centered, draw=black, fill=white, drop shadow]

\tikzstyle{shadow} = [rectangle, align=center, rounded corners, minimum width=1.5cm, minimum height=1.5cm, text centered, draw=gray!30, fill=gray!30]

\tikzstyle{shadow_rect} = [shadow, minimum width=2.7cm, minimum height=1.9cm]

\tikzstyle{action} = [rectangle, align=center, minimum width=3cm, minimum height=1cm, text centered, draw=black, fill=white ]

%% file: Sections/1_introduction.tex
Generative image models such as OpenAI's DALL·E or Stability AI's Stable Diffusion have advanced rapidly and seen a great rise in popularity over the last few years. To train models like these, millions of images are needed leading to the requirement of huge image datasets. 

This need has led to image generation models being trained on images without the needed consent or necessary permissions. As a consequence - besides the violation of the ownership of images - image generation models have been able to copy the style of artists and generate images in the likeness of others without the permission to do so. Such infringements affecting both individuals and organisations can be difficult to prove, as it requires knowledge of the images used to train the generative model.

The aim of this paper and research is to investigate \acrlong{MIA}s (MIA) on \acrlong{LDM}s (LDM) \cite{rombach2022highresolution}. To scope the project only Stable Diffusion v1.5 \cite{rombach2022highresolution} fine-tuned on face images is considered. Being able to infer if an image was part of a dataset used for training a generative model would considerably help individuals and organisations who suspect that their images have been unrightfully used. As mentioned in \cite{Dubinski2023TowardsMR}, evaluating MIA against a fine-tuned model is a potential pitfall\footnote{The reason it is seen as a pitfall is that fine-tuning a model easily leads to over-fitting to an image dataset resulting in higher accuracy on predicting member images. This is also shown in \cite{Carlini2023ExtractingTD}.}. This paper intentionally focuses on fine-tuned models as they are commonly used in real-life applications. It should be kept in mind that the results presented do not apply to non-fine-tuned models. 

\subsection{Definition of the Target model} \label{sec:intro_def_target_model}
In this paper, 'Target Model' $\mathbf{M_T}$  will be used to denote the model which the MIA is performed against. The model $\mathbf{M_T}$ is in this paper characterised by its ability to turn text, $\mathcal{T}$, into some $H\times W$-dimensional image with 3 RGB colour channels, i.e. $\mathbb{R}^{(H,W,3)}$.
\begin{equation}
    \mathbf{M_T} : \mathcal{T} \rightarrow \mathbb{R}^{(H,W,3)}
\end{equation}
This is the model for which it is desired to infer whether an image belongs to its training set $\mathbf{D_{Target}}$. Throughout the paper, only a black-box setup will be considered. This means that the target model can only be used as intended, i.e. providing textual input to generate output. No additional knowledge of training data, model weights, and etc. is available.

\input{Figures/adversial_finetuning_model}
The Target model $\mathbf{M_T}$ could be produced as shown on \cref{fig:adversial_finetuning_model} where a LDM (such as Stable Diffusion) is fine-tuned on a dataset to produce very domain-specific images which imitate the dataset. This could be images that are in the likeness of a specific artist's style or a group of people.

The Adversarial LDM will have obtained its high quality generation abilities from the pre-training on some image dataset (such as LAION-5B) making it able to generalise and mix styles with the images from $\mathbf{D_{Target}}$. It could be difficult to infer membership when a large image generation model contains information from hundreds of millions of images. However, as the model fine-tunes on a smaller set of images, $\mathbf{D_{Target}}$, it would enhance the information leakage from $\mathbf{D_{Target}}$ and it could be possible to infer whether the images have been used in the fine-tuning or not. 

\subsection{Definition of the Attack model}\label{sec:intro_def_attack_model}
The 'Attack Model' $\mathbf{M_A}$  denotes the model trained to infer the membership of a query image in relation to the target model's training set. It is characterised by taking some image $\mathbb{R}^{(H,W,3)}$ and translating it to a value $\mathcal{P}$ between 0 and 1 expressing the predicted probability of the image being part of $\mathbf{D_{Target}}$.

\begin{equation}
    \mathbf{M_A} : \mathbb{R}^{(H,W,3)} \rightarrow \mathcal{P}
\end{equation}

The attack model will be trained using supervised learning. To create the training set for the attack model, the training positives will be obtained by using the target model to generate images. This is done under the assumption that the target model's output leaks information of the training data. This assumption is required as the goal is to make the attack model learn to recognise the data distribution of the target model's training set and not its generated output.

\input{Figures/basic_attack_model_training}

The training negatives are obtained from an 'auxiliary dataset'. The purpose of the images in this dataset is to represent the same or a similar domain as the one the target model has been trained on. The auxiliary dataset should only contain images that the target model has not trained on, so it can be used as training negatives for the attack model. The training process of the attack model can be seen in Figure \ref{fig:basic_attack_model_training}.

%% file: Figures/adversial_finetuning_model.tex
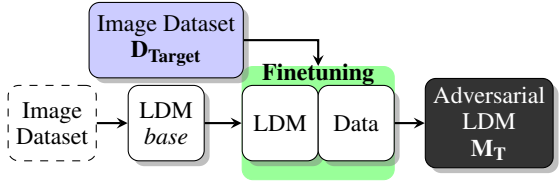
\begin{figure}[h!t]
\small
    \centering
    \begin{tikzpicture}[node distance=2.0cm]
    \node (og_image_dataset) [detail] {Image\\Dataset};
    \node (ldm) [normal, right of=og_image_dataset, xshift=-0.5cm] {LDM\\\textit{base}};
    \node (finetuning) [shadow_rect, right of=ldm, mygreen, minimum height=1.5cm, minimum width=2cm] {};
     \node[action] [above of=finetuning, yshift=-1.35cm, draw=none, fill=none] {\textbf{Finetuning}};
     \node (adversial_ldm) [black, right of=finetuning, xshift=0.25cm] {Adversarial\\LDM\\$\mathbf{M_T}$};
     \node (unlawful_data) [positive, above of=ldm, yshift=-0.9cm] {Image Dataset\\$\mathbf{D_{Target}}$};
     \node (ldm2) [normal, right of=ldm, xshift=-0.5cm] {LDM};
     \node [normal, right of=ldm2, xshift=-1.0cm] {Data};

    \draw [arrow] (og_image_dataset) -- (ldm);
    \draw [arrow] (ldm) -- (finetuning);
    \draw [arrow] (unlawful_data.east) -- (finetuning.north|-unlawful_data) -- (finetuning);
    \draw [arrow] (finetuning) -- (adversial_ldm);

    \end{tikzpicture}
    \caption{An approach a malicious actor could use to obtain a LDM which is trained on unrightfully obtained data.}
    \label{fig:adversial_finetuning_model}
\end{figure}

%% file: Figures/basic_attack_model_training.tex
\begin{figure}[h!t]
\small
    \centering
    \begin{tikzpicture}[node distance=1.35cm]
    \node (supervised_learning) [shadow] {};
    \node [action, above of=supervised_learning, yshift=-0.7cm, draw=none, fill=none] {\textbf{Supervised Learning}};
    \node (train_data) [normal, left of=supervised_learning, xshift=-0.1cm] {$\mathbf{D_{train}}$};
    \node (aux_data) [negative, yshift=0.15cm, below of=train_data]{Auxiliary Data\\\textit{images}};
    \node (gen_img) [normal, left of=train_data]{Output\\\textit{images}};
    \node (target_model) [black, left of=gen_img] {$\mathbf{M_{T}}$};
    \node (og_img_dataset) [positive, yshift=0.15cm, below of=target_model] {$\mathbf{D_{Target}}$};
    \node (model) [normal, left of=supervised_learning, xshift=1.3cm] {Model};
    \node (attack_model) [attack, right of=supervised_learning] {$\mathbf{M_A}$};

    \draw [arrow] (aux_data) -- (train_data);
    \draw [arrow] (gen_img) -- (train_data);
    \draw[arrow] (og_img_dataset) -- (target_model);
    \draw[arrow] (target_model) -- (gen_img);
    \draw[arrow] (train_data) -- (model);
    \draw[arrow] (model) -- (attack_model);

    \end{tikzpicture}
    \caption{The approach taken to train the Attack Model $\mathbf{M_{A}}$. Images ($\mathbf{D_{Target}}$) are used to train $\mathbf{M_T}$ which then outputs some images which are used as positives in $\mathbf{D_{Train}}$. The negatives in $\mathbf{D_{Train}}$ come from an auxiliary dataset. A model is then trained on $\mathbf{D_{Train}}$ using supervised learning to produce the final attack model $\mathbf{M_A}.$}
    \label{fig:basic_attack_model_training}
\end{figure}
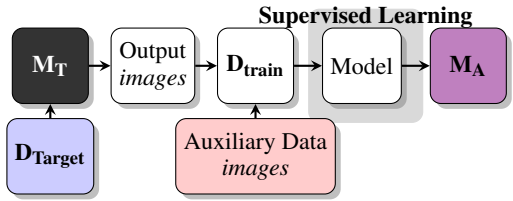

%% file: Sections/2_related_works.tex
In 2017 the paper \textit{Membership Inference Attacks against Machine Learning Models} \cite{shokri2017membership} proposed how to conduct a \acrlong{MIA} against classification models in a black-box setup. The paper investigates the possibility of creating "shadow models" which mimic a target model and uses them to train an attack model for the task of classifying the membership of data samples, i.e. whether they belong to the training set of a target model or not. A shadow model as described in the paper is a model that aims to be similar to the target model, i.e. have a similar architecture, be trained in a similar way, use similar training data etc. The motivation for using shadow models is that it allows for training an attack model using supervised learning.\\ 
This is possible by using the training and test data for the shadow models along with their outputted classification vectors when training the attack models (training set are members, test set are non-members). The paper shows that MIAs can be feasible in a black-box setup.

Besides targeting classification models, multiple papers have also investigated the possibility of performing MIAs on generative models. One such paper is \textit{GAN-Leaks: A Taxonomy of Membership Inference Attacks against Generative Models} \cite{Chen_2020}. The paper has its main focus on MIAs against \acrlong{GAN}s and describes multiple attack scenarios including a full black-box attack. Common for all attack scenarios presented in the paper is the assumption that the probability a generator can produce a sample is proportional to that sample being a member of the generator's training set. This assumption is made as the generator model is trained to approximate the distribution of the training data. The full black-box attack presented works by sampling from a generator and then finding the sample closest to the query sample (ie. the sample to determine the membership of) using a distance metric. The most similar generated sample is then used to approximate the probability that the generator can create the sample which is then used to predict the membership.\\
A generalised approach to MIA was proposed in the paper \textit{Generated Distributions Are All You Need for Membership Inference Attacks Against Generative Models} \cite{zhang2023generated}. The proposed approach does not require shadow models, works in a black-box setup, and can be used against multiple generative models. The core idea for the technique presented is to take advantage of the similar data distribution between the training data of a generative model and its output as done by \cite{Chen_2020}. The similarity between the data distribution relies on the model over-fitting to its training data. This similarity functions as an information leakage, making the generated output describe traits of the model's training data.\\
In the paper, supervised learning is used for training an attack model for the membership inference task. This is done by querying the target model to generate output used as training positives. This is assumed to be representative as training positives due to the assumption of a similar data distribution between the target models training data and its output. The training negatives are obtained from an auxiliary dataset. The paper uses the Resnet-18 architecture for the attack model.

%% file: Sections/3_data.tex
The data used can be divided into two groups. The data used to fine-tune the target model, and the data used to train the attack model.

The focus of this paper is face-images. An image dataset is needed for the experiments and the images should not be contained in LAION-5B\footnote{The images should not be contained in LAION-5B as it was used to train the target model (Stable Diffusion v1.5) and a portion of the image dataset should be used as non-members}. Data from two universities are chosen which have publicly available images of their employees on their websites. The universities are the \acrfull{DTU} and \acrfull{AAU}.

\subsection{Data Source for the Target Model} \label{sec:data_source_for_target_model}
To fine-tune the target model with face-images three different face datasets were considered:
\begin{itemize}
    \item $\mathbf{D_{DTU}}$: Images scraped from DTU orbit.
    \item $\mathbf{D_{AAU}}$: Images scraped from AAU vbn.
    \item $\mathbf{D_{LFW}}$: Labeled Faces in the Wild \cite{LFWTech}. It contains 9,452 images.
\end{itemize}

After collection, the two image data sets $\mathbf{D_{DTU}}$ and $\mathbf{D_{AAU}}$ were partitioned into two subsets, thus producing four different datasets: $\mathbf{D_{DTU}^{seen}}$, $\mathbf{D_{DTU}^{unseen}}$, $\mathbf{D_{AAU}^{seen}}$, and $\mathbf{D_{AAU}^{unseen}}$. The 'seen' datasets are used to fine-tune the target model while the 'unseen' datasets are only used to test/train the attack model. Two more datasets are created which are copied from $\mathbf{D_{DTU}^{seen}}$. One set of the images gets a visible DTU watermark in the top right corner, and the other set gets overlayed with an almost invisible DTU logo, this is shown in \cref{fig:watermark_examples}. On \cref{tab:datasets_description_target} there is a description of all the variations of datasets used to fine-tune the target model.

\input{Tables/datasets_description_target}

\subsection{Data Source for the Attack Model} \label{sec:data_source_for_attack_model}
\begin{figure}[h!t]
\centering
\subfloat[$\mathbf{D_{DTU}^{gen}}$]{\label{attack_model_image_example-a}{\includegraphics[scale=0.065]{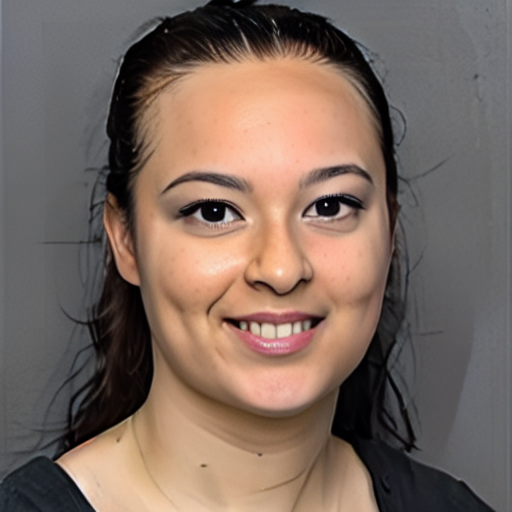}}}%
\hfill
\subfloat[$\mathbf{D_{AAU}^{gen}}$]{\label{attack_model_image_example-b}{\includegraphics[scale=0.065]{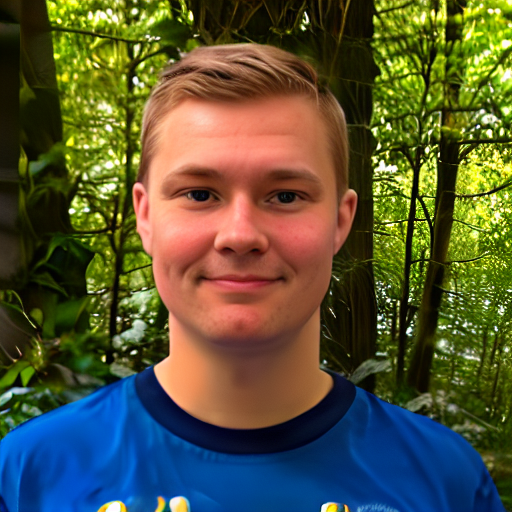}}}
\hfill
\subfloat[\hwmgen]{\label{attack_model_image_example-c}{\includegraphics[scale=0.065]{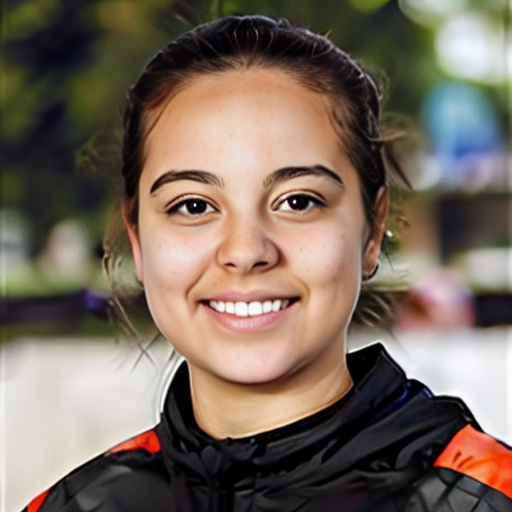}}}
\hfill
\subfloat[\wmgen]{\label{attack_model_image_example-d}{\includegraphics[scale=0.065]{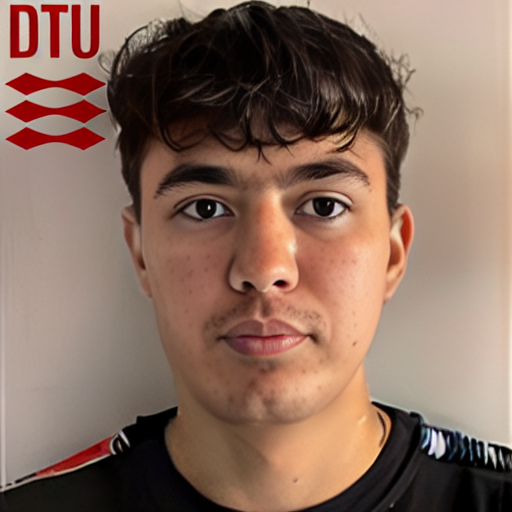}}}
\hfill
\subfloat[$\mathbf{D_{DTU+LFW}^{gen}}$]{\label{attack_model_image_example-e}{\includegraphics[scale=0.065]{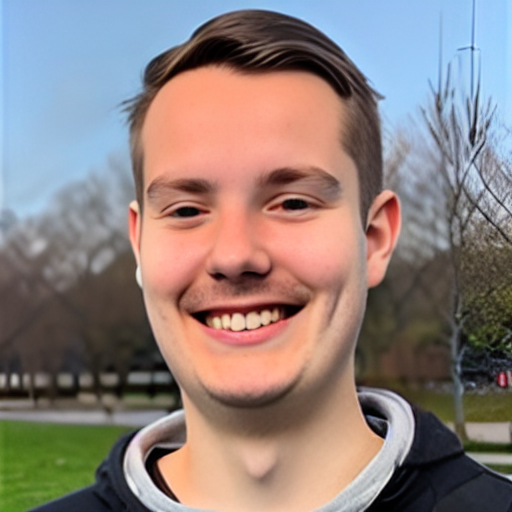}}}
\hfill
\subfloat[$\mathbf{D_{NFT}^{gen}}$]{\label{attack_model_image_example-f}{\includegraphics[scale=0.065]{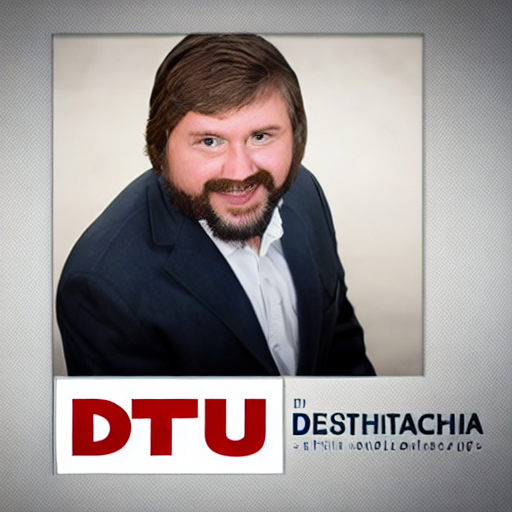}}}
\caption{These are examples of the data in the different generated image datasets which are used to train the attack model.}
\label{fig:attack_model_image_example}
\end{figure}

As mentioned in \cref{sec:data_source_for_target_model} $\mathbf{D_{DTU}^{unseen}}$ and $\mathbf{D_{AAU}^{unseen}}$ are not used to fine-tune the target model, this way they can be used to test or train the attack model with the certainty that there is no data leakage into the generated images. The output of the target models are used in the training of the attack model. On table \cref{tab:datasets_description_attack} there is a description of all the different image-sets used for training / testing the attack model in different experiments. On \cref{fig:attack_model_image_example} examples can be seen of the generated images from the target models.

\input{Tables/datasets_description_attack}

%% file: Tables/datasets_description_target.tex
\bgroup
\def\arraystretch{1.0}
\begin{table*}[h!t]
\small
\centering
\caption{This table describes the different datasets used for fine-tuning the target models.}
\label{tab:datasets_description_target}
\begin{tabular}{c|c|c}
\hline
\textbf{Symbol} & \textbf{Size, n} & \textbf{Description} \\ \hline
$\mathbf{D_{DTU}^{seen}}$ & 1,120 & Partition of $\mathbf{D_{DTU}}$ used to fine-tune the target model $\mathbf{M_T^{DTU}}$ \\ \hline
$\mathbf{D_{AAU}^{seen}}$ & 1,120 & Partition of $\mathbf{D_{AAU}}$ used to fine-tune the target model $\mathbf{M_T^{AAU}}$ \\ \hline
\wmseen & 1,120 & Partition of $\mathbf{D_{DTU}}$ with watermarks used to fine-tune the target model \wmtar \\ \hline
\hwmseen & 1,120 & Partition of $\mathbf{D_{DTU}}$ with hidden watermarks used to fine-tune \hwmtar \\ \hline
$\mathbf{D_{DTU+LFW}}$ & 2,240 & \begin{tabular}[c]{@{}c@{}}A combination of images from $\mathbf{D_{DTU}}$ and $\mathbf{D_{LFW}}$ used to fine-tune $\mathbf{M_T^{DTU+LFW}}$\end{tabular} \\ \hline
\end{tabular}
\end{table*}

%% file: Tables/datasets_description_attack.tex
\bgroup
\def\arraystretch{1.0}
\begin{table*}[h!t]
\small
\centering
\caption{This table describes the different datasets used for fine-tuning the attack models.}
\label{tab:datasets_description_attack}
\begin{tabular}{c|c|c}
\hline
\textbf{Symbol} & \textbf{Size, n} & \textbf{Description} \\ \hline
$\mathbf{D_{DTU}^{unseen}}$ & 1,103 & A partition of $\mathbf{D_{DTU}}$ that is \textbf{not} used to fine-tune any image generation model. \\ \hline
$\mathbf{D_{AAU}^{unseen}}$ & 978 & A partition of $\mathbf{D_{AAU}}$ that is \textbf{not} used to fine-tune any image generation model. \\ \hline
$\mathbf{D_{LFW}}$ & 9,452 & Labeled Faces in the Wild: A Database for Studying Face Recognition \cite{LFWTech} \\ \hline
$\mathbf{D_{DTU}^{gen}}$ & 2,500 & Generated by a LDM which has been fine-tuned on the $\mathbf{D_{DTU}^{seen}}$ image set, i.e. $\mathbf{M_T^{DTU}}$ \\ \hline
$\mathbf{D_{AAU}^{gen}}$ & 2,500 & Generated by a LDM that has been fine-tuned on the $\mathbf{D_{AAU}^{seen}}$ image set, i.e. $\mathbf{M_T^{AAU}}$ \\ \hline
\wmgen & 2,500 & Generated by a LDM which has been fine-tuned on the \wmseen image set, i.e. \wmtar \\ \hline
\hwmgen & 2,500 & Generated by a LDM which has been fine-tuned on the \hwmseen image set, i.e. \hwmtar \\ \hline
$\mathbf{D_{DTU+LFW}^{gen}}$ & 2,500 & Generated by a LDM that has been fine-tuned on the  $\mathbf{D_{DTU+LFW}}$ image set, i.e. $\mathbf{M_T^{DTU+LFW}}$ \\ \hline
$\mathbf{D_{NFT}^{gen}}$ & $2\times$ 2,500 & \begin{tabular}[c]{@{}c@{}}Generated by a Non Fine-Tuned (NFT) target model $\mathbf{M_T}$, \\ i.e. Stable Diffusion out-of-the-box. This dataset also comes in two versions, one which was \\ prompted with \texttt{"a dtu headshot"} and one which was prompted with \texttt{"a aau headshot"}\end{tabular} \\ \hline
\end{tabular}
\end{table*}

%% file: Sections/4_method.tex
\paragraph{The target dataset}
 $\mathbf{D_{target}}$ used for fine-tuning $\mathbf{M_T}$ is a dataset consisting of image-text pairs with the text describing the image. A BLIP model \cite{blip}  is used to auto-label the images. When $\mathbf{D_{DTU}^{seen}}$ is used to create $\mathbf{D_{target}}$, the labels generated with BLIP are conditioned with \texttt{"a dtu headshot of a"} and for $\mathbf{D_{AAU}^{seen}}$, \texttt{"a aau headshot of a"} is used. In other words, all labels in $\mathbf{D_{target}^{DTU}}$ begin with \texttt{"a dtu headshot of a (...)"}.
 
\paragraph{The Training Positives} 
$\mathbf{D^{gen}}$ are generated using the fine-tuned model $\mathbf{M_T}$. 2,500 images are generated using 100 inference steps and a guidance scale of 7.5.  When using the model fine-tuned on DTU images, the prompt \texttt{"a dtu headshot"} is used and for the model fine-tuned on AAU images, the prompt \texttt{"a aau headshot"} is used. 25 images are generated per seed.

\paragraph{The Auxiliary Data}
i.e. the training negatives for the supervised learning of $\mathbf{M_A}$ are supposed to represent all images from the same (or a similar) domain as $\mathbf{D_{target}}$ not seen by $\mathbf{M_T}$. The auxiliary data must not have been used for training $\mathbf{M_T}$, i.e. the data should not be part of $\mathbf{D_{target}}$ or the original training data for $\mathbf{M_T}$. In this paper the domain of interest is face images, so the auxiliary data should be face images not seen by $\mathbf{M_T}$.\\
 In \cite{zhang2023generated} they propose using an auxiliary dataset consisting of real images from other datasets. The experiments conducted in this paper will default to constructing the auxiliary datasets using another generative model to generate images. This is done to ensure the attack model does not simply learn to classify real images and generated images. This is also mentioned as the 2nd pitfall for MIA on LDM's in \cite{Dubinski2023TowardsMR}. It should be noted that this is more computational expensive than simply using some dataset, as it will require training a generative model when performing a MIA. 

\subsection{Model Specifications}
\paragraph{The Target Model} 
$\mathbf{M_T}$ used for this project is a fine-tuned version of Stable Diffusion v1.5 \cite{rombach2022highresolution}. Stable Diffusion v1.5 is trained on a subset of LAION-5B. The model is then fine-tuned on each of the datasets presented in \cref{sec:data_source_for_target_model} resulting in multiple models, e.g. $\mathbf{M_T^{DTU}}$ which is the SD v1.5 model fine-tuned on $\mathbf{D_{DTU}^{seen}}$ or $\mathbf{M_T^{DTU \ WM}}$ which is the SD v1.5 model fine-tuned on \wmseen instead.

\paragraph{The Attack Model} 
used is Resnet-18 which was introduced in 2015 \cite{DBLP:journals/corr/HeZRS15}. The pretrained weights\footnote{The weights are available here \url{https://download.pytorch.org/models/resnet18-f37072fd.pth}} are kept, however a fully connected layer with two neurons replaces the standard 1000-neuron final layer. The loss function used is ca\-te\-go\-ri\-cal cross\-en\-tro\-py and the Adam optimizer is used for fast con\-ver\-gen\-ce.

\subsection{Experimental Setup} \label{sec:Test_Setup}
Multiple experiments are performed to determine the success and performance of MIAs against the target model, $\mathbf{M_T}$. Similar for all tests is that they aim to help investigate the possibility of a successful MIA on a $\mathbf{M_T}$ in the setup depicted in Figure \ref{fig:our_setup_model}. As illustrated in the figure, some $\mathbf{M_T}$ is being fine-tuned on an image dataset which has been unrightfully obtained, $\mathbf{D_{target}}$.

\input{Figures/our_setup_model}

The target model $\mathbf{M_T}$ is then queried to generate output images used as training positives in $\mathbf{D_{train}}$ and some auxiliary data is added to $\mathbf{D_{train}}$ as negatives. The attack model $\mathbf{M_A}$ is then trained on $\mathbf{D_{train}}$ using supervised learning. The resulting model $\mathbf{M_A}$ is then queried on images to classify whether they were part of $\mathbf{D_{target}}$ or not. 15\% of the test data is used for validation of $\mathbf{M_A}$. 

For the experiments  $\mathbf{M_T}$ is fine-tuned using a known $\mathbf{D_{target}}$, hereby the ground-truth is known when creating $\mathbf{D_{train}}$ and querying the resulting model $\mathbf{M_A}$. This makes it possible to determine the performance of $\mathbf{M_A}$ and the MIA. On \cref{fig:testing_attack_model} the approach to testing the attack model is shown. 

\input{Figures/testing_attack_model}

%% file: Figures/our_setup_model.tex
\begin{figure*}[h!t]
\smaller
    \centering
    \begin{tikzpicture}[node distance=1.2cm]
    \node (supervised_learning) [shadow] {};
    \node [action][above of=supervised_learning, yshift=-0.38cm, draw=none, fill=none]{\textbf{Supervised}\\$\mathbf{Learning}$};
    \node (train_data) [normal, left of=supervised_learning, xshift=-1cm] {Training Data\\$\mathbf{D_{train}}$};
    \node (aux_data) [negative, left of=train_data, yshift=-1.25cm, xshift=-2.0cm]{Auxiliary Images\\$\mathbf{-}$};
    \node (gen_img) [normal, above of=aux_data]{Output Images\\$\mathbf{+}$};
    \node [normal, left of=supervised_learning, xshift=1.2cm] {Model};
    \node (attack_model) [attack, right of=supervised_learning, xshift=0.5cm] {$\mathbf{M_A}$};

    \node (og_image_dataset) [detail, above of=gen_img, xshift=-5cm] {Image\\Dataset};
    \node (finetuning) [shadow_rect, right of=og_image_dataset, mygreen, minimum height=1.5cm, minimum width=2cm, xshift=1cm] {};
     \node[action] [above of=finetuning, yshift=-0.6cm, draw=none, fill=none] {\textbf{Finetuning}};
     \node (adversial_ldm) [black, right of=finetuning, xshift=1.5cm] {Adversarial LDM\\$\mathbf{M_T}$};
     \node (unlawful_data) [positive, below of=og_image_dataset, yshift=0cm] {Image Dataset\\$\mathbf{D_{Target}}$};
     \node (ldm2) [normal, right of=og_image_dataset, xshift=0.5cm] {LDM};
     \node [normal, right of=ldm2, xshift=-0.25cm] {Data};
    \node (query) [normal, above of=attack_model] {Query\\$\mathbf{Q}$};
    \node (output) [normal, below of=attack_model] {$\mathbf{+} \ \texttt{if} \ \mathbf{Q} \in \mathbf{D_{Target}}$\\$\mathcal{-} \ \texttt{if} \ \mathbf{Q}  \notin \mathbf{D_{Target}}$};

    \draw [arrow] (og_image_dataset) -- (finetuning);
    \draw [arrow] (unlawful_data.east) -- (finetuning.north|-unlawful_data) -- (finetuning);
    \draw [arrow] (finetuning) -- (adversial_ldm);

    \draw [arrow] (aux_data) -- (train_data.south|-aux_data) -- (train_data);
    \draw [arrow] (gen_img) -- (adversial_ldm.east|-gen_img) --  (adversial_ldm.east|-train_data)  -- (train_data);
    \draw[arrow] (train_data) -- (supervised_learning);
    \draw[arrow] (supervised_learning) -- (attack_model);
    \draw[arrow] (adversial_ldm) -- (gen_img);
    \draw [arrow] (query) -- (attack_model);
    \draw [arrow] (attack_model) -- (output);

    \end{tikzpicture}
    \caption{A graphic representation of how an attack model $\mathbf{M_A}$ could be built and later queried with an image $\mathbf{Q}$.}
    \label{fig:our_setup_model}
\end{figure*}
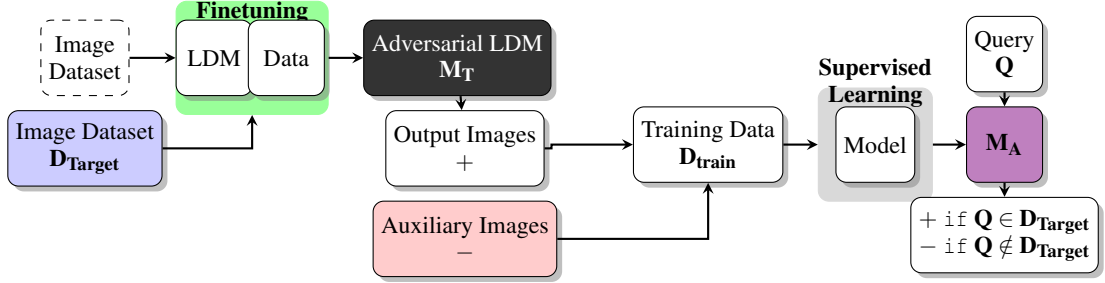

%% file: Figures/testing_attack_model.tex
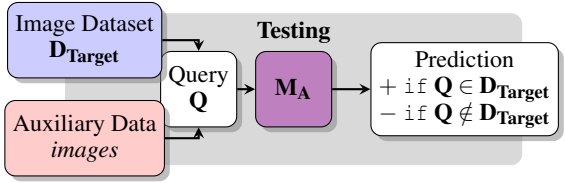
\begin{figure}[h!t]
\small
    \centering
    \begin{tikzpicture}[node distance=1.0cm]

    \node (testing) [shadow_rect, minimum height=2cm, minimum width=6cm] {};
    \node[action] [above of=testing, yshift=-0.25cm, draw=none, fill=none] {\textbf{Testing}};
    \node (attack_model) [attack] {$\mathbf{M_A}$};
    \node (test) [normal, left of=attack_model, xshift=-0.25cm] {Query\\$\mathbf{Q}$};
    \node (unlawful_data) [positive, left of=test, yshift=0.65cm, xshift=-0.5cm] {Image Dataset\\$\mathbf{D_{Target}}$};
    \node (aux_data) [negative, left of=test, yshift=-0.65cm, xshift=-0.5cm]{Auxiliary Data\\\textit{images}};
    \node (output) [normal, right of=attack_model, xshift=1.25cm] {Prediction\\$\mathbf{+} \ \texttt{if} \ \mathbf{Q} \in \mathbf{D_{Target}}$\\$\mathcal{-} \ \texttt{if} \ \mathbf{Q}  \notin \mathbf{D_{Target}}$};

    \draw [arrow] (attack_model) -- (output);
    \draw [arrow] (test) -- (attack_model);
    \draw [arrow] (unlawful_data) -- (test.north|-unlawful_data) -- (test);
    \draw [arrow] (aux_data) -- (test.south|-aux_data) -- (test);
    \end{tikzpicture}
    \caption{$\mathbf{M_{A}}$ is given a set $\mathbf{Q}$ of images combined from $\mathbf{D_{Target}}$ and an auxiliary image data set. This allows for evaluating the performance of $\mathbf{M_{A}}$}
    \label{fig:testing_attack_model}
\end{figure}

%% file: Sections/5_results.tex
For each test, the attack model has been trained 5 different times with 5 different seeds to calculate the 95\% confidence intervals of the metrics of interest. Zero-shot classification using the CLIP model \cite{radford2021learning} is used as a baseline. On \cref{tab:results_table}, all the results are summarised. Note that the motivation/basis of several tests are different, therefore they should not all be directly compared on only the AUC score.

\input{Tables/results_table}

\paragraph{The Impact of the Relationship between training and test negatives} can be seen on \cref{tab:results_table} No. 3, 4, and 7. The MIA is successful across all three tests: $\mathbf{D_{DTU}^{seen}}$ vs $\mathbf{D_{AAU}^{seen}}$, $\mathbf{D_{DTU}^{seen}}$ vs $\mathbf{D_{AAU}^{unseen}}$, and $\mathbf{D_{DTU}^{seen}}$ vs $\mathbf{D_{LFW}}$. All three test uses $\mathbf{D_{DTU}^{gen}}$ and $\mathbf{D_{AAU}^{gen}}$ for the training of $\mathbf{M_A}$. No significance difference was found in the MIA performance for the different settings of test positives and negatives. In our case it does not seem to matter if the train negatives were generated using the test negatives (cf. \cref{tab:results_table} No. 3 vs 4). Neither did it seem to matter if the training and test negatives are sampled from the same data distribution distinct from the training positives or not (cf. \cref{tab:results_table} No. 3 vs 7). $\mathbf{M_A}$ clearly outperforms the baseline in test 3 and 4, but the baseline model performs better on experiment 7 with a near perfect AUC for the baseline. This indicates that using a not-generated image-set against a generated image-set artificially boosts the MIA performance.

\paragraph{The effect of using real images for the auxiliary dataset} can be seen on test (\cref{tab:results_table} No. 2). It uses 2,000 images from $\mathbf{D_{DTU}^{gen}}$ and 2,000 images from $\mathbf{D_{AAU}}$ to train $\mathbf{M_A}$. The attack model is then tested on 952 $\mathbf{D_{DTU}^{seen}}$ images (positives) and 8,034 $\mathbf{D_{LFW}}$ images (negatives). This was done to test the effect of using real images for the auxiliary set (training negatives). As can be seen on \cref{tab:results_table} experiment No. 2, $\mathbf{M_A}$ is successful in the test and achieves an AUC of $\sim0.71 \pm0.02$.\\
When comparing this with \cref{tab:results_table} No. 7, which shows the same test except $\mathbf{M_A}$ is trained using the generated auxiliary set $\mathbf{D_{AAU}^{gen}}$ instead, it becomes apparent that it is beneficial for the attack model to use a generated auxiliary set as it's training negatives. Zhang et al. finds that using a generated auxiliary dataset is comparable to using real images \cite{zhang2023generated}. In our case, we find that using a generated auxiliary dataset is significantly better than using real images.

It can also be seen on \cref{tab:results_table} No. 2, that the baseline model excels in classifying $\mathbf{D_{DTU}^{seen}}$ and $\mathbf{D_{LFW}}$ as members and non-members and clearly outperforms $\mathbf{M_A}$ for this test with a near perfect AUC\footnote{This test is identical to the test depicted on \cref{tab:results_table} No. 7, as only the training negatives differ in the two tests, and the training negatives are not considered by the baseline.}. Note that the perfect AUC is likely due to the non-generated auxiliary data being too dissimilar to the positives, which artificially boosts the MIA performance.

\paragraph{Results of MIA without fine-tuning }
can be seen on Test no. 10. It is carried out where $\mathbf{D_{NFT}^{gen}}$ is used as both the training positives and negatives for  $\mathbf{M_{A}}$. As to stay consistent with the other tests, $\mathbf{M_T}$ is prompted with \texttt{"a dtu headshot"} and \texttt{"a aau headshot"} to generate two different $\mathbf{D_{NFT}^{gen}}$. As no information leakage can exist between neither the positives or negatives (unless the pretraining of SD 1.5 includes images from AAU or DTU), the MIA should not perform better than random guessing. However it did, which could be explained by a coincidence in data distribution similarity.

For the Test no. 9, $\mathbf{M_A}$ was first trained on $\mathbf{D_{NFT}^{gen}}$ as positives and $\mathbf{D_{AAU}^{gen}}$ as negatives. Then it was tested using $\mathbf{D_{DTU}^{seen}}$ as positives and $\mathbf{D_{AAU}^{unseen}}$ as negatives.
The MIA was still successful although the AUC score achieved by $\mathbf{M_A}$, $0.66\pm0.03$, is significantly worse than in all other tests on $\mathbf{D_{DTU}^{seen}}$ vs $\mathbf{D_{AAU}^{unseen}}$.

As there is no relation between training and test positives, this result shows that there is information leakage between training negatives $\mathbf{D_{AAU}^{gen}}$ and test negatives $\mathbf{D_{AAU}^{unseen}}$. The implication of this discovery is that the tests using $\mathbf{D_{AAU}^{gen}}$ as training negatives and $\mathbf{D_{AAU}^{unseen}}$ (or $\mathbf{D_{AAU}^{seen}}$) as test negatives have a artificial boost in their performance and thus inflated metrics.  This could be explained by the fact that they originate from the same data distribution.

\paragraph{Relationship between training time of target model and success of MIA.}
The MIA performance against target models fine-tuned on $\mathbf{D_{target}^{DTU}}$ for an increasing number of epochs can be seen on \cref{tab:results_table} experiment No. 4. For all tests, $\mathbf{M_A}$ is trained on $\mathbf{D_{DTU}^{gen}}$ and $\mathbf{D_{AAU}^{gen}}$ and tested on $\mathbf{D_{DTU}^{seen}}$ and $\mathbf{D_{AAU}^{unseen}}$. For 400 epochs, the AUC score is significantly better than the one for 100 epochs (the same can not be concluded for 400 vs 50 epochs, as the intervals barely overlap). As seen on the 50 epoch experiment, the MIA is still successful when $\mathbf{M_T^{DTU}}$ has trained for 50 epochs and results in comparable performance to the tests against target models trained for 400 epochs. The baseline appears to perform equally well across the different training times and thus does not appear to gain extended knowledge of the underlying training data distribution with increasing training time.

\paragraph{A mix of DTU and LFW in the target dataset.}
Here $\mathbf{M_A}$ is trained using $\mathbf{D_{AAU}^{gen}}$ as negatives and a mix of DTU and LFW as positives: $\mathbf{D_{DTU+LFW}^{gen}}$ (balanced training with $2,500$ images from each set). Note that only $\mathbf{D_{DTU}^{seen}}$ are test positives. The result shown on \cref{tab:results_table} experiment No. 13 shows that $\mathbf{M_A}$ still performs a successful MIA even if only half of the data used to fine-tune $\mathbf{M_T}$ is of interest. It is not significantly worse than when the target model is only fine-tuned on $\mathbf{D_{DTU}^{seen}}$, which was the case in experiment No. 4 ($0.83\pm0.02$ vs $0.86\pm0.02$). While much worse than $\mathbf{M_A}$, the baseline model is also able to perform a successful MIA in this test with close to the same performance when compared to test No. 4.

\begin{figure}[h!t]
\centering
\subfloat[$s=0$]{\label{cfg-0-example} {\includegraphics[scale=0.075]{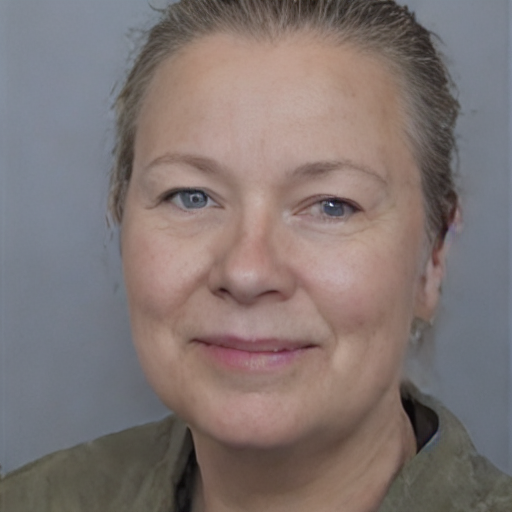}}}
\hfill
\subfloat[$s=4$]{\label{cfg-4-example} {\includegraphics[scale=0.075]{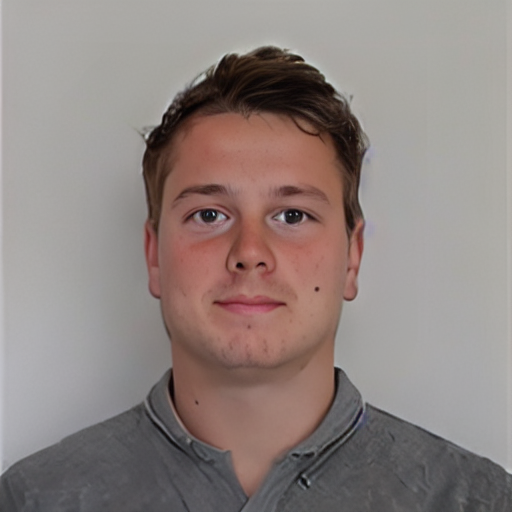}}}%
\hfill
\subfloat[$s=8$]{\label{cfg-8-example} {\includegraphics[scale=0.075]{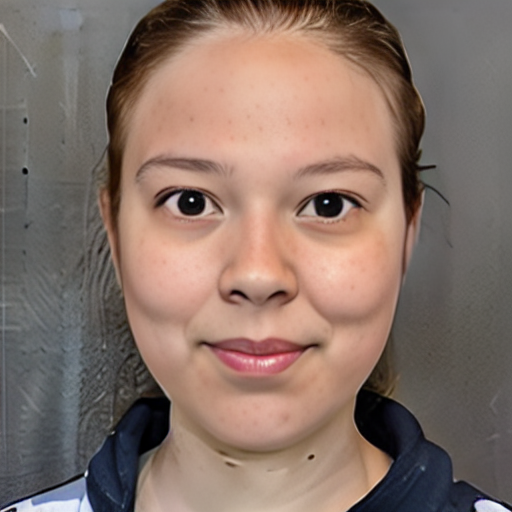}}}%
\hfill
\subfloat[$s=12$]{\label{cfg-12-example} {\includegraphics[scale=0.075]{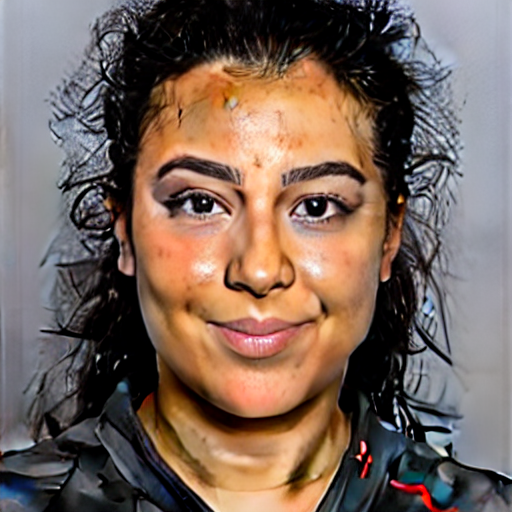}}}%
\hfill
\subfloat[$s=16$]{\label{cfg-16-example} {\includegraphics[scale=0.075]{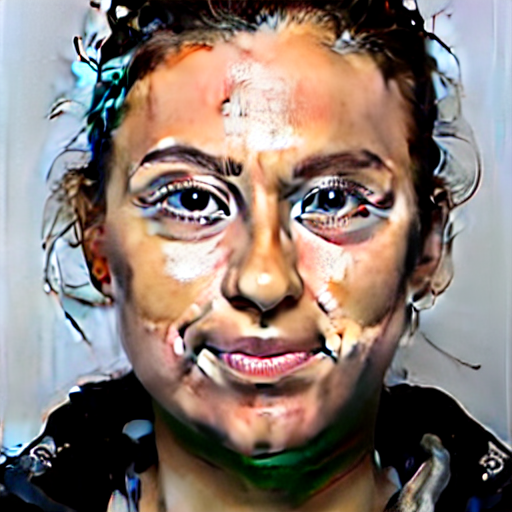}}}%
\caption{Examples of images generated by $\mathbf{M_T^{DTU}}$ using different guidance scales.}
\label{fig:guidance_scale_examples}
\end{figure}

\paragraph{A different prompt for target model inference}
has influence on the generated output. The result of using the prompt \texttt{"a profile picture"} instead of \texttt{"a dtu headshot"} for inference to generate $\mathbf{D_{DTU}^{gen}}$ used as training positives for $\mathbf{M_A}$ is shown on \cref{tab:results_table} test No. 5. The performance of the attack model is not significantly different when conditioning the target model with \texttt{"a profile picture"} for inference compared to \cref{tab:results_table} test No. 4 ($0.82\pm0.05$ vs $0.86\pm0.02$). This could indicate that the generated distribution still contains the features which allow the Resnet-18 to make good predictions.  An explanation could be that the 400 epochs of fine-tuning on the "dtu" label and images has made the latent space more uniform. This is supported by the fact that on \cref{fig:guidance_scale_examples} the image with guidance scale $s=0$ still produces a headshot, even though according to \cref{eq:cfg} it is unconditioned.

\begin{equation}\label{eq:cfg}
\epsilon_t=\epsilon_{t,uncond} + s \cdot (\epsilon_{t,\tau(y)} - \epsilon_{t,uncond})    
\end{equation}

\paragraph{Recognising seen vs unseen samples from the same data distribution.}

The MIA on $\mathbf{D_{DTU}^{seen}}$ vs $\mathbf{D_{DTU}^{unseen}}$ performed poorly as seen on \cref{tab:results_table} test No. 8. This shows that the MIA presented in this paper does not seem successful in the task of identifying membership on an individual basis (for a single data-point), but instead is viable for the task of inferring the membership of a dataset as a whole. These results indicate some sort of shared characteristic among the collection of images. The nature of this shared characteristic is unknown. It is a reasonable assumption that images taken at the same locations/universities/organisations share some features.

\begin{figure}[h!t]
\centering
\subfloat[visible watermark]{\label{wm-example} {\includegraphics[scale=0.1]{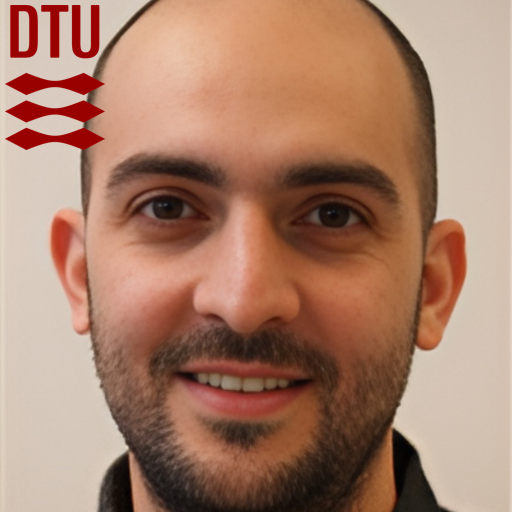}}}
\hfill
\subfloat[25\% visible]{\label{hwm-25-example} {\includegraphics[scale=0.1]{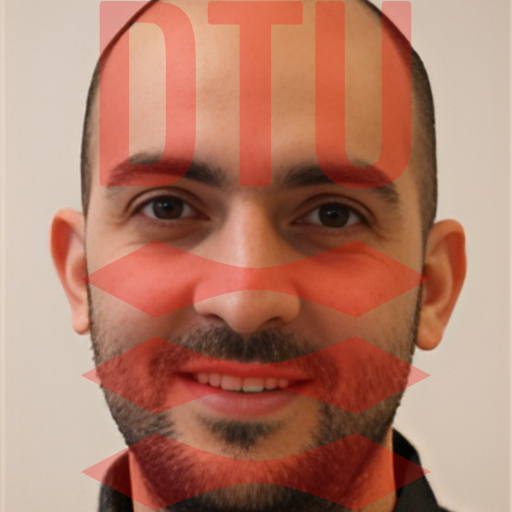}}}%
\hfill
\subfloat[1\% visible]{\label{hwm-1-example} {\includegraphics[scale=0.1]{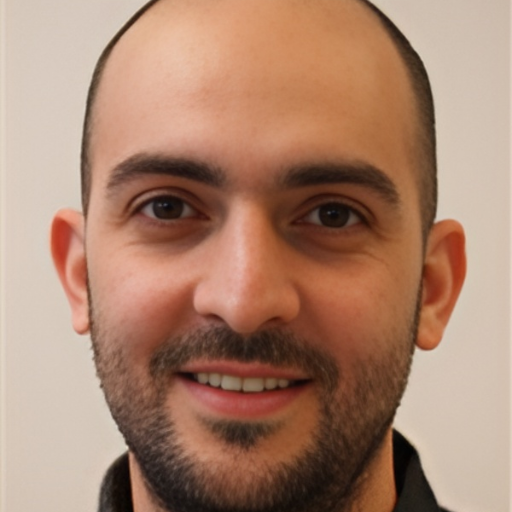}}}%
\caption{(\protect\subref{wm-example}) shows an example of the visible watermark. (\protect\subref{hwm-25-example}) illustrates the shape of the hidden watermark. (\protect\subref{hwm-1-example}) shows the actual hidden watermark used for testing.}
\label{fig:watermark_examples}
\end{figure}

\paragraph{Using watermarks for MIA enhancement}
can be seen on \cref{tab:results_table} test No. 11 and 12. The visible watermark tested in \wmseen vs $\mathbf{D_{AAU}^{unseen}}$ were very effective and lead to a near perfect classification of the test set by the attack model. The use of a hidden watermark however did not show any improvement compared to using no watermarks (cf. \cref{tab:results_table} test No. 12 vs No. 4) which shows that either the target model did not learn to mimic the hidden watermark, or that the attack model was not able to pick up on the nearly invisible watermark.
An example of the watermarks is shown on \cref{fig:watermark_examples}.

\begin{figure}[h!t]
    \centering
    \scalebox{0.25}{\input{Graphs/guidance_scale.pgf}}
    \caption{AUC scores for the test of $\mathbf{M_A}$ where the training positives have been generated with different guidance scales. The FID score compared to the original DTU data is also shown, a higher AUC score correlates with a lower FID score. This is consistent with \cite{Carlini2023ExtractingTD}.}
    \label{fig:guidance_auc}
\end{figure}
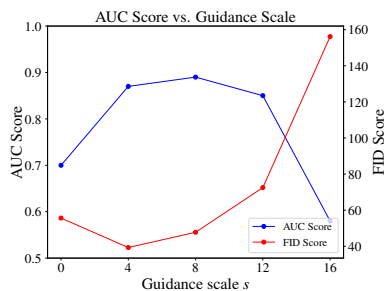

\paragraph{MIA Performance for Different Guidance Scales} can be seen on \cref{fig:guidance_auc}, the performance of a MIA in our case is sensitive to the guidance scale used when generating training positives. The AUC score achieved by $\mathbf{M_A}$ is highest when the guidance scale is between 4 and 12. Looking at \cref{fig:guidance_scale_examples}, we see that even with a guidance scale of $s=0$, $\mathbf{M_T^{DTU}}$ still generates a headshot, despite having no guidance of what to generate. This indicates that fine-tuning $\mathbf{M_T^{DTU}}$ for 400 epochs has introduced enough bias that it assumes noise to stem from images of headshots. For $s=16$ visible artifacts appear, distorting the face (\cref{cfg-16-example}) which might also explain the drop in AUC score. It is also notable that for $s=0$, $\mathbf{M_T^{DTU}}$ achieves an AUC of $\sim0.7$ even though it is trained on positives generated without any guidance, which demonstrates that the target model $\mathbf{M_T^{DTU}}$ trained on 400 epochs leaks information of its training data even when not conditioned on a prompt.

\makeatletter
\setlength{\@fptop}{0pt}
\setlength{\@fpbot}{0pt plus 1fil}
\makeatother

%% file: Tables/results_table.tex
\bgroup
\def\arraystretch{1.0}
\begin{table*}[h!t]
\centering
\begin{threeparttable}
\caption{This table contains a summary of all results from all tests. It also includes training and test datasets. The interval on the Resnet-18 AUC Score is the 95\%-confidence interval calculated over 5 repetitions.}
\label{tab:results_table}
\smaller
\begin{tabular}{l|c|cccc|cc}
\hline
{No.} & {Experiment} & {Train Pos.} & {Train Neg.} & {Test Pos.} & {Test Neg.} & {R18 AUC} & {CLIP AUC} \\ \hline
1&{\begin{tabular}[c]{@{}l@{}}Generated DTU vs\\ generated AAU\end{tabular}} & $\mathbf{D_{DTU}^{gen}}$ & $\mathbf{D_{AAU}^{gen}}$ & $\mathbf{D_{DTU}^{gen}}$ & $\mathbf{D_{AAU}^{gen}}$ & $1.00\pm0.00$ & 0.94 \\ \hline
2&{\begin{tabular}[c]{@{}l@{}}Generated DTU vs\\ non-generated AAU\end{tabular}} & $\mathbf{D_{DTU}^{gen}}$ & $\mathbf{D_{AAU}}$ & $\mathbf{D_{DTU}^{seen}}$ & $\mathbf{D_{LFW}}$ & $0.71\pm0.02$ & 0.99 \\ \hline
3&{DTU vs AAU Seen} & $\mathbf{D_{DTU}^{gen}}$ & $\mathbf{D_{AAU}^{gen}}$ & $\mathbf{D_{DTU}^{seen}}$ & $\mathbf{D_{AAU}^{seen}}$ & $0.86\pm0.01$ & 0.63 \\ \hline
4&{\begin{tabular}[c]{@{}l@{}}DTU vs AAU Unseen\\ Trained 50, 100,\\  and 400 Epochs\end{tabular}} & \multirow{3}{*}{$\mathbf{D_{DTU}^{gen}}$} & \multirow{3}{*}{$\mathbf{D_{AAU}^{gen}}$} & \multirow{3}{*}{$\mathbf{D_{DTU}^{seen}}$} & \multirow{3}{*}{$\mathbf{D_{AAU}^{unseen}}$} & \begin{tabular}[c]{@{}c@{}}$0.86\pm0.02$,\\ $0.82\pm0.01$,\\ $0.86\pm0.02$\end{tabular} & \begin{tabular}[c]{@{}c@{}}0.63,\\ 0.62,\\ 0.63\end{tabular} \\ \cline{1-2} \cline{7-8} 
5& {Generalised Prompt} &  &  &  &  & $0.82\pm0.05$ & 0.56 \\ \cline{1-2} \cline{7-8} 
6& {\begin{tabular}[c]{@{}l@{}}Guidance Scale\\ s = 0, 4, 8, 12, 16\end{tabular}} &  &  &  &  & \begin{tabular}[c]{@{}c@{}}0.70, 0.87, 0.89,\\  0.85, 0.58\end{tabular} & - \\ \hline
7& {DTU vs LFW} & $\mathbf{D_{DTU}^{gen}}$ & $\mathbf{D_{AAU}^{gen}}$ & $\mathbf{D_{DTU}^{seen}}$ & $\mathbf{D_{LFW}}$ & $0.89\pm0.04$ & 0.99 \\ \hline
8& {DTU vs DTU} & $\mathbf{D_{DTU}^{gen}}$ & $\mathbf{D_{AAU}^{gen}}$ & $\mathbf{D_{DTU}^{seen}}$ & $\mathbf{D_{DTU}^{unseen}}$ & $0.53\pm0.00$ & 0.52 \\ \hline
9& {NFT vs AAU} & $\mathbf{D_{NFT}^{gen}}$ & $\mathbf{D_{AAU}^{gen}}$ & $\mathbf{D_{DTU}^{seen}}$ & $\mathbf{D_{AAU}^{unseen}}$ & $0.66\pm0.03$ & 0.55 \\ \hline
10& {NFT vs NFT} & $\mathbf{D_{NFT}^{gen}}$ & $\mathbf{D_{NFT}^{gen}}$ & $\mathbf{D_{DTU}^{seen}}$ & $\mathbf{D_{AAU}^{unseen}}$ & $0.54\pm0.03$ & 0.55 \\ \hline
11& {Watermark} & \wmgen & $\mathbf{D_{AAU}^{gen}}$ & \wmseen & $\mathbf{D_{AAU}^{unseen}}$ & $1.00\pm0.00$ & 0.95 \\ \hline
12& {\begin{tabular}[c]{@{}l@{}}Hidden Watermark\end{tabular}} & \hwmgen & $\mathbf{D_{AAU}^{gen}}$ & \hwmseen & $\mathbf{D_{AAU}^{unseen}}$ & $0.83\pm0.03$ & 0.57 \\ \hline
13& {DTU+LFW vs AAU} & $\mathbf{D_{DTU+LFW}^{gen}}$ & $\mathbf{D_{AAU}^{gen}}$ & $\mathbf{D_{DTU}^{seen}}$ & $\mathbf{D_{AAU}^{unseen}}$ & $0.83\pm0.02$ & 0.64 \\ \hline
\end{tabular}%
\end{threeparttable}
\end{table*}

%% file: Graphs/guidance_scale.pgf
\begingroup%
\makeatletter%
\begin{pgfpicture}%
\pgfpathrectangle{\pgfpointorigin}{\pgfqpoint{8.000000in}{6.000000in}}%
\pgfusepath{use as bounding box, clip}%
\begin{pgfscope}%
\pgfsetbuttcap%
\pgfsetmiterjoin%
\definecolor{currentfill}{rgb}{1.000000,1.000000,1.000000}%
\pgfsetfillcolor{currentfill}%
\pgfsetlinewidth{0.000000pt}%
\definecolor{currentstroke}{rgb}{1.000000,1.000000,1.000000}%
\pgfsetstrokecolor{currentstroke}%
\pgfsetdash{}{0pt}%
\pgfpathmoveto{\pgfqpoint{0.000000in}{0.000000in}}%
\pgfpathlineto{\pgfqpoint{8.000000in}{0.000000in}}%
\pgfpathlineto{\pgfqpoint{8.000000in}{6.000000in}}%
\pgfpathlineto{\pgfqpoint{0.000000in}{6.000000in}}%
\pgfpathlineto{\pgfqpoint{0.000000in}{0.000000in}}%
\pgfpathclose%
\pgfusepath{fill}%
\end{pgfscope}%
\begin{pgfscope}%
\pgfsetbuttcap%
\pgfsetmiterjoin%
\definecolor{currentfill}{rgb}{1.000000,1.000000,1.000000}%
\pgfsetfillcolor{currentfill}%
\pgfsetlinewidth{0.000000pt}%
\definecolor{currentstroke}{rgb}{0.000000,0.000000,0.000000}%
\pgfsetstrokecolor{currentstroke}%
\pgfsetstrokeopacity{0.000000}%
\pgfsetdash{}{0pt}%
\pgfpathmoveto{\pgfqpoint{0.893781in}{0.789058in}}%
\pgfpathlineto{\pgfqpoint{7.027052in}{0.789058in}}%
\pgfpathlineto{\pgfqpoint{7.027052in}{5.633997in}}%
\pgfpathlineto{\pgfqpoint{0.893781in}{5.633997in}}%
\pgfpathlineto{\pgfqpoint{0.893781in}{0.789058in}}%
\pgfpathclose%
\pgfusepath{fill}%
\end{pgfscope}%
\begin{pgfscope}%
\pgfsetbuttcap%
\pgfsetroundjoin%
\definecolor{currentfill}{rgb}{0.000000,0.000000,0.000000}%
\pgfsetfillcolor{currentfill}%
\pgfsetlinewidth{0.803000pt}%
\definecolor{currentstroke}{rgb}{0.000000,0.000000,0.000000}%
\pgfsetstrokecolor{currentstroke}%
\pgfsetdash{}{0pt}%
\pgfsys@defobject{currentmarker}{\pgfqpoint{0.000000in}{-0.048611in}}{\pgfqpoint{0.000000in}{0.000000in}}{%
\pgfpathmoveto{\pgfqpoint{0.000000in}{0.000000in}}%
\pgfpathlineto{\pgfqpoint{0.000000in}{-0.048611in}}%
\pgfusepath{stroke,fill}%
}%
\begin{pgfscope}%
\pgfsys@transformshift{1.172566in}{0.789058in}%
\pgfsys@useobject{currentmarker}{}%
\end{pgfscope}%
\end{pgfscope}%
\begin{pgfscope}%
\definecolor{textcolor}{rgb}{0.000000,0.000000,0.000000}%
\pgfsetstrokecolor{textcolor}%
\pgfsetfillcolor{textcolor}%
\pgftext[x=1.172566in,y=0.691836in,,top]{\color{textcolor}{\rmfamily\fontsize{20.000000}{24.000000}\selectfont\catcode`\^=\active\def^{\ifmmode\sp\else\^{}\fi}\catcode`\%=\active\def
\end{pgfscope}%
\begin{pgfscope}%
\pgfsetbuttcap%
\pgfsetroundjoin%
\definecolor{currentfill}{rgb}{0.000000,0.000000,0.000000}%
\pgfsetfillcolor{currentfill}%
\pgfsetlinewidth{0.803000pt}%
\definecolor{currentstroke}{rgb}{0.000000,0.000000,0.000000}%
\pgfsetstrokecolor{currentstroke}%
\pgfsetdash{}{0pt}%
\pgfsys@defobject{currentmarker}{\pgfqpoint{0.000000in}{-0.048611in}}{\pgfqpoint{0.000000in}{0.000000in}}{%
\pgfpathmoveto{\pgfqpoint{0.000000in}{0.000000in}}%
\pgfpathlineto{\pgfqpoint{0.000000in}{-0.048611in}}%
\pgfusepath{stroke,fill}%
}%
\begin{pgfscope}%
\pgfsys@transformshift{2.566491in}{0.789058in}%
\pgfsys@useobject{currentmarker}{}%
\end{pgfscope}%
\end{pgfscope}%
\begin{pgfscope}%
\definecolor{textcolor}{rgb}{0.000000,0.000000,0.000000}%
\pgfsetstrokecolor{textcolor}%
\pgfsetfillcolor{textcolor}%
\pgftext[x=2.566491in,y=0.691836in,,top]{\color{textcolor}{\rmfamily\fontsize{20.000000}{24.000000}\selectfont\catcode`\^=\active\def^{\ifmmode\sp\else\^{}\fi}\catcode`\%=\active\def
\end{pgfscope}%
\begin{pgfscope}%
\pgfsetbuttcap%
\pgfsetroundjoin%
\definecolor{currentfill}{rgb}{0.000000,0.000000,0.000000}%
\pgfsetfillcolor{currentfill}%
\pgfsetlinewidth{0.803000pt}%
\definecolor{currentstroke}{rgb}{0.000000,0.000000,0.000000}%
\pgfsetstrokecolor{currentstroke}%
\pgfsetdash{}{0pt}%
\pgfsys@defobject{currentmarker}{\pgfqpoint{0.000000in}{-0.048611in}}{\pgfqpoint{0.000000in}{0.000000in}}{%
\pgfpathmoveto{\pgfqpoint{0.000000in}{0.000000in}}%
\pgfpathlineto{\pgfqpoint{0.000000in}{-0.048611in}}%
\pgfusepath{stroke,fill}%
}%
\begin{pgfscope}%
\pgfsys@transformshift{3.960417in}{0.789058in}%
\pgfsys@useobject{currentmarker}{}%
\end{pgfscope}%
\end{pgfscope}%
\begin{pgfscope}%
\definecolor{textcolor}{rgb}{0.000000,0.000000,0.000000}%
\pgfsetstrokecolor{textcolor}%
\pgfsetfillcolor{textcolor}%
\pgftext[x=3.960417in,y=0.691836in,,top]{\color{textcolor}{\rmfamily\fontsize{20.000000}{24.000000}\selectfont\catcode`\^=\active\def^{\ifmmode\sp\else\^{}\fi}\catcode`\%=\active\def
\end{pgfscope}%
\begin{pgfscope}%
\pgfsetbuttcap%
\pgfsetroundjoin%
\definecolor{currentfill}{rgb}{0.000000,0.000000,0.000000}%
\pgfsetfillcolor{currentfill}%
\pgfsetlinewidth{0.803000pt}%
\definecolor{currentstroke}{rgb}{0.000000,0.000000,0.000000}%
\pgfsetstrokecolor{currentstroke}%
\pgfsetdash{}{0pt}%
\pgfsys@defobject{currentmarker}{\pgfqpoint{0.000000in}{-0.048611in}}{\pgfqpoint{0.000000in}{0.000000in}}{%
\pgfpathmoveto{\pgfqpoint{0.000000in}{0.000000in}}%
\pgfpathlineto{\pgfqpoint{0.000000in}{-0.048611in}}%
\pgfusepath{stroke,fill}%
}%
\begin{pgfscope}%
\pgfsys@transformshift{5.354342in}{0.789058in}%
\pgfsys@useobject{currentmarker}{}%
\end{pgfscope}%
\end{pgfscope}%
\begin{pgfscope}%
\definecolor{textcolor}{rgb}{0.000000,0.000000,0.000000}%
\pgfsetstrokecolor{textcolor}%
\pgfsetfillcolor{textcolor}%
\pgftext[x=5.354342in,y=0.691836in,,top]{\color{textcolor}{\rmfamily\fontsize{20.000000}{24.000000}\selectfont\catcode`\^=\active\def^{\ifmmode\sp\else\^{}\fi}\catcode`\%=\active\def
\end{pgfscope}%
\begin{pgfscope}%
\pgfsetbuttcap%
\pgfsetroundjoin%
\definecolor{currentfill}{rgb}{0.000000,0.000000,0.000000}%
\pgfsetfillcolor{currentfill}%
\pgfsetlinewidth{0.803000pt}%
\definecolor{currentstroke}{rgb}{0.000000,0.000000,0.000000}%
\pgfsetstrokecolor{currentstroke}%
\pgfsetdash{}{0pt}%
\pgfsys@defobject{currentmarker}{\pgfqpoint{0.000000in}{-0.048611in}}{\pgfqpoint{0.000000in}{0.000000in}}{%
\pgfpathmoveto{\pgfqpoint{0.000000in}{0.000000in}}%
\pgfpathlineto{\pgfqpoint{0.000000in}{-0.048611in}}%
\pgfusepath{stroke,fill}%
}%
\begin{pgfscope}%
\pgfsys@transformshift{6.748267in}{0.789058in}%
\pgfsys@useobject{currentmarker}{}%
\end{pgfscope}%
\end{pgfscope}%
\begin{pgfscope}%
\definecolor{textcolor}{rgb}{0.000000,0.000000,0.000000}%
\pgfsetstrokecolor{textcolor}%
\pgfsetfillcolor{textcolor}%
\pgftext[x=6.748267in,y=0.691836in,,top]{\color{textcolor}{\rmfamily\fontsize{20.000000}{24.000000}\selectfont\catcode`\^=\active\def^{\ifmmode\sp\else\^{}\fi}\catcode`\%=\active\def
\end{pgfscope}%
\begin{pgfscope}%
\definecolor{textcolor}{rgb}{0.000000,0.000000,0.000000}%
\pgfsetstrokecolor{textcolor}%
\pgfsetfillcolor{textcolor}%
\pgftext[x=3.960417in,y=0.366003in,,top]{\color{textcolor}{\rmfamily\fontsize{24.000000}{28.800000}\selectfont\catcode`\^=\active\def^{\ifmmode\sp\else\^{}\fi}\catcode`\%=\active\def
\end{pgfscope}%
\begin{pgfscope}%
\pgfsetbuttcap%
\pgfsetroundjoin%
\definecolor{currentfill}{rgb}{0.000000,0.000000,0.000000}%
\pgfsetfillcolor{currentfill}%
\pgfsetlinewidth{0.803000pt}%
\definecolor{currentstroke}{rgb}{0.000000,0.000000,0.000000}%
\pgfsetstrokecolor{currentstroke}%
\pgfsetdash{}{0pt}%
\pgfsys@defobject{currentmarker}{\pgfqpoint{-0.048611in}{0.000000in}}{\pgfqpoint{-0.000000in}{0.000000in}}{%
\pgfpathmoveto{\pgfqpoint{-0.000000in}{0.000000in}}%
\pgfpathlineto{\pgfqpoint{-0.048611in}{0.000000in}}%
\pgfusepath{stroke,fill}%
}%
\begin{pgfscope}%
\pgfsys@transformshift{0.893781in}{0.789058in}%
\pgfsys@useobject{currentmarker}{}%
\end{pgfscope}%
\end{pgfscope}%
\begin{pgfscope}%
\definecolor{textcolor}{rgb}{0.000000,0.000000,0.000000}%
\pgfsetstrokecolor{textcolor}%
\pgfsetfillcolor{textcolor}%
\pgftext[x=0.421559in, y=0.689336in, left, base]{\color{textcolor}{\rmfamily\fontsize{20.000000}{24.000000}\selectfont\catcode`\^=\active\def^{\ifmmode\sp\else\^{}\fi}\catcode`\%=\active\def
\end{pgfscope}%
\begin{pgfscope}%
\pgfsetbuttcap%
\pgfsetroundjoin%
\definecolor{currentfill}{rgb}{0.000000,0.000000,0.000000}%
\pgfsetfillcolor{currentfill}%
\pgfsetlinewidth{0.803000pt}%
\definecolor{currentstroke}{rgb}{0.000000,0.000000,0.000000}%
\pgfsetstrokecolor{currentstroke}%
\pgfsetdash{}{0pt}%
\pgfsys@defobject{currentmarker}{\pgfqpoint{-0.048611in}{0.000000in}}{\pgfqpoint{-0.000000in}{0.000000in}}{%
\pgfpathmoveto{\pgfqpoint{-0.000000in}{0.000000in}}%
\pgfpathlineto{\pgfqpoint{-0.048611in}{0.000000in}}%
\pgfusepath{stroke,fill}%
}%
\begin{pgfscope}%
\pgfsys@transformshift{0.893781in}{1.758046in}%
\pgfsys@useobject{currentmarker}{}%
\end{pgfscope}%
\end{pgfscope}%
\begin{pgfscope}%
\definecolor{textcolor}{rgb}{0.000000,0.000000,0.000000}%
\pgfsetstrokecolor{textcolor}%
\pgfsetfillcolor{textcolor}%
\pgftext[x=0.421559in, y=1.658324in, left, base]{\color{textcolor}{\rmfamily\fontsize{20.000000}{24.000000}\selectfont\catcode`\^=\active\def^{\ifmmode\sp\else\^{}\fi}\catcode`\%=\active\def
\end{pgfscope}%
\begin{pgfscope}%
\pgfsetbuttcap%
\pgfsetroundjoin%
\definecolor{currentfill}{rgb}{0.000000,0.000000,0.000000}%
\pgfsetfillcolor{currentfill}%
\pgfsetlinewidth{0.803000pt}%
\definecolor{currentstroke}{rgb}{0.000000,0.000000,0.000000}%
\pgfsetstrokecolor{currentstroke}%
\pgfsetdash{}{0pt}%
\pgfsys@defobject{currentmarker}{\pgfqpoint{-0.048611in}{0.000000in}}{\pgfqpoint{-0.000000in}{0.000000in}}{%
\pgfpathmoveto{\pgfqpoint{-0.000000in}{0.000000in}}%
\pgfpathlineto{\pgfqpoint{-0.048611in}{0.000000in}}%
\pgfusepath{stroke,fill}%
}%
\begin{pgfscope}%
\pgfsys@transformshift{0.893781in}{2.727034in}%
\pgfsys@useobject{currentmarker}{}%
\end{pgfscope}%
\end{pgfscope}%
\begin{pgfscope}%
\definecolor{textcolor}{rgb}{0.000000,0.000000,0.000000}%
\pgfsetstrokecolor{textcolor}%
\pgfsetfillcolor{textcolor}%
\pgftext[x=0.421559in, y=2.627312in, left, base]{\color{textcolor}{\rmfamily\fontsize{20.000000}{24.000000}\selectfont\catcode`\^=\active\def^{\ifmmode\sp\else\^{}\fi}\catcode`\%=\active\def
\end{pgfscope}%
\begin{pgfscope}%
\pgfsetbuttcap%
\pgfsetroundjoin%
\definecolor{currentfill}{rgb}{0.000000,0.000000,0.000000}%
\pgfsetfillcolor{currentfill}%
\pgfsetlinewidth{0.803000pt}%
\definecolor{currentstroke}{rgb}{0.000000,0.000000,0.000000}%
\pgfsetstrokecolor{currentstroke}%
\pgfsetdash{}{0pt}%
\pgfsys@defobject{currentmarker}{\pgfqpoint{-0.048611in}{0.000000in}}{\pgfqpoint{-0.000000in}{0.000000in}}{%
\pgfpathmoveto{\pgfqpoint{-0.000000in}{0.000000in}}%
\pgfpathlineto{\pgfqpoint{-0.048611in}{0.000000in}}%
\pgfusepath{stroke,fill}%
}%
\begin{pgfscope}%
\pgfsys@transformshift{0.893781in}{3.696022in}%
\pgfsys@useobject{currentmarker}{}%
\end{pgfscope}%
\end{pgfscope}%
\begin{pgfscope}%
\definecolor{textcolor}{rgb}{0.000000,0.000000,0.000000}%
\pgfsetstrokecolor{textcolor}%
\pgfsetfillcolor{textcolor}%
\pgftext[x=0.421559in, y=3.596299in, left, base]{\color{textcolor}{\rmfamily\fontsize{20.000000}{24.000000}\selectfont\catcode`\^=\active\def^{\ifmmode\sp\else\^{}\fi}\catcode`\%=\active\def
\end{pgfscope}%
\begin{pgfscope}%
\pgfsetbuttcap%
\pgfsetroundjoin%
\definecolor{currentfill}{rgb}{0.000000,0.000000,0.000000}%
\pgfsetfillcolor{currentfill}%
\pgfsetlinewidth{0.803000pt}%
\definecolor{currentstroke}{rgb}{0.000000,0.000000,0.000000}%
\pgfsetstrokecolor{currentstroke}%
\pgfsetdash{}{0pt}%
\pgfsys@defobject{currentmarker}{\pgfqpoint{-0.048611in}{0.000000in}}{\pgfqpoint{-0.000000in}{0.000000in}}{%
\pgfpathmoveto{\pgfqpoint{-0.000000in}{0.000000in}}%
\pgfpathlineto{\pgfqpoint{-0.048611in}{0.000000in}}%
\pgfusepath{stroke,fill}%
}%
\begin{pgfscope}%
\pgfsys@transformshift{0.893781in}{4.665009in}%
\pgfsys@useobject{currentmarker}{}%
\end{pgfscope}%
\end{pgfscope}%
\begin{pgfscope}%
\definecolor{textcolor}{rgb}{0.000000,0.000000,0.000000}%
\pgfsetstrokecolor{textcolor}%
\pgfsetfillcolor{textcolor}%
\pgftext[x=0.421559in, y=4.565287in, left, base]{\color{textcolor}{\rmfamily\fontsize{20.000000}{24.000000}\selectfont\catcode`\^=\active\def^{\ifmmode\sp\else\^{}\fi}\catcode`\%=\active\def
\end{pgfscope}%
\begin{pgfscope}%
\pgfsetbuttcap%
\pgfsetroundjoin%
\definecolor{currentfill}{rgb}{0.000000,0.000000,0.000000}%
\pgfsetfillcolor{currentfill}%
\pgfsetlinewidth{0.803000pt}%
\definecolor{currentstroke}{rgb}{0.000000,0.000000,0.000000}%
\pgfsetstrokecolor{currentstroke}%
\pgfsetdash{}{0pt}%
\pgfsys@defobject{currentmarker}{\pgfqpoint{-0.048611in}{0.000000in}}{\pgfqpoint{-0.000000in}{0.000000in}}{%
\pgfpathmoveto{\pgfqpoint{-0.000000in}{0.000000in}}%
\pgfpathlineto{\pgfqpoint{-0.048611in}{0.000000in}}%
\pgfusepath{stroke,fill}%
}%
\begin{pgfscope}%
\pgfsys@transformshift{0.893781in}{5.633997in}%
\pgfsys@useobject{currentmarker}{}%
\end{pgfscope}%
\end{pgfscope}%
\begin{pgfscope}%
\definecolor{textcolor}{rgb}{0.000000,0.000000,0.000000}%
\pgfsetstrokecolor{textcolor}%
\pgfsetfillcolor{textcolor}%
\pgftext[x=0.421559in, y=5.534275in, left, base]{\color{textcolor}{\rmfamily\fontsize{20.000000}{24.000000}\selectfont\catcode`\^=\active\def^{\ifmmode\sp\else\^{}\fi}\catcode`\%=\active\def
\end{pgfscope}%
\begin{pgfscope}%
\definecolor{textcolor}{rgb}{0.000000,0.000000,0.000000}%
\pgfsetstrokecolor{textcolor}%
\pgfsetfillcolor{textcolor}%
\pgftext[x=0.366003in,y=3.211528in,,bottom,rotate=90.000000]{\color{textcolor}{\rmfamily\fontsize{24.000000}{28.800000}\selectfont\catcode`\^=\active\def^{\ifmmode\sp\else\^{}\fi}\catcode`\%=\active\def
\end{pgfscope}%
\begin{pgfscope}%
\pgfpathrectangle{\pgfqpoint{0.893781in}{0.789058in}}{\pgfqpoint{6.133272in}{4.844938in}}%
\pgfusepath{clip}%
\pgfsetrectcap%
\pgfsetroundjoin%
\pgfsetlinewidth{1.505625pt}%
\definecolor{currentstroke}{rgb}{0.000000,0.000000,1.000000}%
\pgfsetstrokecolor{currentstroke}%
\pgfsetdash{}{0pt}%
\pgfpathmoveto{\pgfqpoint{1.172566in}{2.727034in}}%
\pgfpathlineto{\pgfqpoint{2.566491in}{4.374313in}}%
\pgfpathlineto{\pgfqpoint{3.960417in}{4.568110in}}%
\pgfpathlineto{\pgfqpoint{5.354342in}{4.180515in}}%
\pgfpathlineto{\pgfqpoint{6.748267in}{1.564249in}}%
\pgfusepath{stroke}%
\end{pgfscope}%
\begin{pgfscope}%
\pgfpathrectangle{\pgfqpoint{0.893781in}{0.789058in}}{\pgfqpoint{6.133272in}{4.844938in}}%
\pgfusepath{clip}%
\pgfsetbuttcap%
\pgfsetroundjoin%
\definecolor{currentfill}{rgb}{0.000000,0.000000,1.000000}%
\pgfsetfillcolor{currentfill}%
\pgfsetlinewidth{1.003750pt}%
\definecolor{currentstroke}{rgb}{0.000000,0.000000,1.000000}%
\pgfsetstrokecolor{currentstroke}%
\pgfsetdash{}{0pt}%
\pgfsys@defobject{currentmarker}{\pgfqpoint{-0.041667in}{-0.041667in}}{\pgfqpoint{0.041667in}{0.041667in}}{%
\pgfpathmoveto{\pgfqpoint{0.000000in}{-0.041667in}}%
\pgfpathcurveto{\pgfqpoint{0.011050in}{-0.041667in}}{\pgfqpoint{0.021649in}{-0.037276in}}{\pgfqpoint{0.029463in}{-0.029463in}}%
\pgfpathcurveto{\pgfqpoint{0.037276in}{-0.021649in}}{\pgfqpoint{0.041667in}{-0.011050in}}{\pgfqpoint{0.041667in}{0.000000in}}%
\pgfpathcurveto{\pgfqpoint{0.041667in}{0.011050in}}{\pgfqpoint{0.037276in}{0.021649in}}{\pgfqpoint{0.029463in}{0.029463in}}%
\pgfpathcurveto{\pgfqpoint{0.021649in}{0.037276in}}{\pgfqpoint{0.011050in}{0.041667in}}{\pgfqpoint{0.000000in}{0.041667in}}%
\pgfpathcurveto{\pgfqpoint{-0.011050in}{0.041667in}}{\pgfqpoint{-0.021649in}{0.037276in}}{\pgfqpoint{-0.029463in}{0.029463in}}%
\pgfpathcurveto{\pgfqpoint{-0.037276in}{0.021649in}}{\pgfqpoint{-0.041667in}{0.011050in}}{\pgfqpoint{-0.041667in}{0.000000in}}%
\pgfpathcurveto{\pgfqpoint{-0.041667in}{-0.011050in}}{\pgfqpoint{-0.037276in}{-0.021649in}}{\pgfqpoint{-0.029463in}{-0.029463in}}%
\pgfpathcurveto{\pgfqpoint{-0.021649in}{-0.037276in}}{\pgfqpoint{-0.011050in}{-0.041667in}}{\pgfqpoint{0.000000in}{-0.041667in}}%
\pgfpathlineto{\pgfqpoint{0.000000in}{-0.041667in}}%
\pgfpathclose%
\pgfusepath{stroke,fill}%
}%
\begin{pgfscope}%
\pgfsys@transformshift{1.172566in}{2.727034in}%
\pgfsys@useobject{currentmarker}{}%
\end{pgfscope}%
\begin{pgfscope}%
\pgfsys@transformshift{2.566491in}{4.374313in}%
\pgfsys@useobject{currentmarker}{}%
\end{pgfscope}%
\begin{pgfscope}%
\pgfsys@transformshift{3.960417in}{4.568110in}%
\pgfsys@useobject{currentmarker}{}%
\end{pgfscope}%
\begin{pgfscope}%
\pgfsys@transformshift{5.354342in}{4.180515in}%
\pgfsys@useobject{currentmarker}{}%
\end{pgfscope}%
\begin{pgfscope}%
\pgfsys@transformshift{6.748267in}{1.564249in}%
\pgfsys@useobject{currentmarker}{}%
\end{pgfscope}%
\end{pgfscope}%
\begin{pgfscope}%
\pgfsetrectcap%
\pgfsetmiterjoin%
\pgfsetlinewidth{0.803000pt}%
\definecolor{currentstroke}{rgb}{0.000000,0.000000,0.000000}%
\pgfsetstrokecolor{currentstroke}%
\pgfsetdash{}{0pt}%
\pgfpathmoveto{\pgfqpoint{0.893781in}{0.789058in}}%
\pgfpathlineto{\pgfqpoint{0.893781in}{5.633997in}}%
\pgfusepath{stroke}%
\end{pgfscope}%
\begin{pgfscope}%
\pgfsetrectcap%
\pgfsetmiterjoin%
\pgfsetlinewidth{0.803000pt}%
\definecolor{currentstroke}{rgb}{0.000000,0.000000,0.000000}%
\pgfsetstrokecolor{currentstroke}%
\pgfsetdash{}{0pt}%
\pgfpathmoveto{\pgfqpoint{7.027052in}{0.789058in}}%
\pgfpathlineto{\pgfqpoint{7.027052in}{5.633997in}}%
\pgfusepath{stroke}%
\end{pgfscope}%
\begin{pgfscope}%
\pgfsetrectcap%
\pgfsetmiterjoin%
\pgfsetlinewidth{0.803000pt}%
\definecolor{currentstroke}{rgb}{0.000000,0.000000,0.000000}%
\pgfsetstrokecolor{currentstroke}%
\pgfsetdash{}{0pt}%
\pgfpathmoveto{\pgfqpoint{0.893781in}{0.789058in}}%
\pgfpathlineto{\pgfqpoint{7.027053in}{0.789058in}}%
\pgfusepath{stroke}%
\end{pgfscope}%
\begin{pgfscope}%
\pgfsetrectcap%
\pgfsetmiterjoin%
\pgfsetlinewidth{0.803000pt}%
\definecolor{currentstroke}{rgb}{0.000000,0.000000,0.000000}%
\pgfsetstrokecolor{currentstroke}%
\pgfsetdash{}{0pt}%
\pgfpathmoveto{\pgfqpoint{0.893781in}{5.633997in}}%
\pgfpathlineto{\pgfqpoint{7.027053in}{5.633997in}}%
\pgfusepath{stroke}%
\end{pgfscope}%
\begin{pgfscope}%
\definecolor{textcolor}{rgb}{0.000000,0.000000,0.000000}%
\pgfsetstrokecolor{textcolor}%
\pgfsetfillcolor{textcolor}%
\pgftext[x=3.960417in,y=5.718997in,,base]{\color{textcolor}{\rmfamily\fontsize{24.000000}{28.800000}\selectfont\catcode`\^=\active\def^{\ifmmode\sp\else\^{}\fi}\catcode`\%=\active\def
\end{pgfscope}%
\begin{pgfscope}%
\pgfsetbuttcap%
\pgfsetmiterjoin%
\definecolor{currentfill}{rgb}{1.000000,1.000000,1.000000}%
\pgfsetfillcolor{currentfill}%
\pgfsetfillopacity{0.800000}%
\pgfsetlinewidth{1.003750pt}%
\definecolor{currentstroke}{rgb}{0.800000,0.800000,0.800000}%
\pgfsetstrokecolor{currentstroke}%
\pgfsetstrokeopacity{0.800000}%
\pgfsetdash{}{0pt}%
\pgfpathmoveto{\pgfqpoint{5.107497in}{0.900170in}}%
\pgfpathlineto{\pgfqpoint{6.871497in}{0.900170in}}%
\pgfpathquadraticcurveto{\pgfqpoint{6.915941in}{0.900170in}}{\pgfqpoint{6.915941in}{0.944614in}}%
\pgfpathlineto{\pgfqpoint{6.915941in}{1.577058in}}%
\pgfpathquadraticcurveto{\pgfqpoint{6.915941in}{1.621503in}}{\pgfqpoint{6.871497in}{1.621503in}}%
\pgfpathlineto{\pgfqpoint{5.107497in}{1.621503in}}%
\pgfpathquadraticcurveto{\pgfqpoint{5.063053in}{1.621503in}}{\pgfqpoint{5.063053in}{1.577058in}}%
\pgfpathlineto{\pgfqpoint{5.063053in}{0.944614in}}%
\pgfpathquadraticcurveto{\pgfqpoint{5.063053in}{0.900170in}}{\pgfqpoint{5.107497in}{0.900170in}}%
\pgfpathlineto{\pgfqpoint{5.107497in}{0.900170in}}%
\pgfpathclose%
\pgfusepath{stroke,fill}%
\end{pgfscope}%
\begin{pgfscope}%
\pgfsetrectcap%
\pgfsetroundjoin%
\pgfsetlinewidth{1.505625pt}%
\definecolor{currentstroke}{rgb}{0.000000,0.000000,1.000000}%
\pgfsetstrokecolor{currentstroke}%
\pgfsetdash{}{0pt}%
\pgfpathmoveto{\pgfqpoint{5.151941in}{1.450836in}}%
\pgfpathlineto{\pgfqpoint{5.374164in}{1.450836in}}%
\pgfpathlineto{\pgfqpoint{5.596386in}{1.450836in}}%
\pgfusepath{stroke}%
\end{pgfscope}%
\begin{pgfscope}%
\pgfsetbuttcap%
\pgfsetroundjoin%
\definecolor{currentfill}{rgb}{0.000000,0.000000,1.000000}%
\pgfsetfillcolor{currentfill}%
\pgfsetlinewidth{1.003750pt}%
\definecolor{currentstroke}{rgb}{0.000000,0.000000,1.000000}%
\pgfsetstrokecolor{currentstroke}%
\pgfsetdash{}{0pt}%
\pgfsys@defobject{currentmarker}{\pgfqpoint{-0.041667in}{-0.041667in}}{\pgfqpoint{0.041667in}{0.041667in}}{%
\pgfpathmoveto{\pgfqpoint{0.000000in}{-0.041667in}}%
\pgfpathcurveto{\pgfqpoint{0.011050in}{-0.041667in}}{\pgfqpoint{0.021649in}{-0.037276in}}{\pgfqpoint{0.029463in}{-0.029463in}}%
\pgfpathcurveto{\pgfqpoint{0.037276in}{-0.021649in}}{\pgfqpoint{0.041667in}{-0.011050in}}{\pgfqpoint{0.041667in}{0.000000in}}%
\pgfpathcurveto{\pgfqpoint{0.041667in}{0.011050in}}{\pgfqpoint{0.037276in}{0.021649in}}{\pgfqpoint{0.029463in}{0.029463in}}%
\pgfpathcurveto{\pgfqpoint{0.021649in}{0.037276in}}{\pgfqpoint{0.011050in}{0.041667in}}{\pgfqpoint{0.000000in}{0.041667in}}%
\pgfpathcurveto{\pgfqpoint{-0.011050in}{0.041667in}}{\pgfqpoint{-0.021649in}{0.037276in}}{\pgfqpoint{-0.029463in}{0.029463in}}%
\pgfpathcurveto{\pgfqpoint{-0.037276in}{0.021649in}}{\pgfqpoint{-0.041667in}{0.011050in}}{\pgfqpoint{-0.041667in}{0.000000in}}%
\pgfpathcurveto{\pgfqpoint{-0.041667in}{-0.011050in}}{\pgfqpoint{-0.037276in}{-0.021649in}}{\pgfqpoint{-0.029463in}{-0.029463in}}%
\pgfpathcurveto{\pgfqpoint{-0.021649in}{-0.037276in}}{\pgfqpoint{-0.011050in}{-0.041667in}}{\pgfqpoint{0.000000in}{-0.041667in}}%
\pgfpathlineto{\pgfqpoint{0.000000in}{-0.041667in}}%
\pgfpathclose%
\pgfusepath{stroke,fill}%
}%
\begin{pgfscope}%
\pgfsys@transformshift{5.374164in}{1.450836in}%
\pgfsys@useobject{currentmarker}{}%
\end{pgfscope}%
\end{pgfscope}%
\begin{pgfscope}%
\definecolor{textcolor}{rgb}{0.000000,0.000000,0.000000}%
\pgfsetstrokecolor{textcolor}%
\pgfsetfillcolor{textcolor}%
\pgftext[x=5.774164in,y=1.373058in,left,base]{\color{textcolor}{\rmfamily\fontsize{16.000000}{19.200000}\selectfont\catcode`\^=\active\def^{\ifmmode\sp\else\^{}\fi}\catcode`\%=\active\def
\end{pgfscope}%
\begin{pgfscope}%
\pgfsetrectcap%
\pgfsetroundjoin%
\pgfsetlinewidth{1.505625pt}%
\definecolor{currentstroke}{rgb}{1.000000,0.000000,0.000000}%
\pgfsetstrokecolor{currentstroke}%
\pgfsetdash{}{0pt}%
\pgfpathmoveto{\pgfqpoint{5.151941in}{1.123503in}}%
\pgfpathlineto{\pgfqpoint{5.374164in}{1.123503in}}%
\pgfpathlineto{\pgfqpoint{5.596386in}{1.123503in}}%
\pgfusepath{stroke}%
\end{pgfscope}%
\begin{pgfscope}%
\pgfsetbuttcap%
\pgfsetroundjoin%
\definecolor{currentfill}{rgb}{1.000000,0.000000,0.000000}%
\pgfsetfillcolor{currentfill}%
\pgfsetlinewidth{1.003750pt}%
\definecolor{currentstroke}{rgb}{1.000000,0.000000,0.000000}%
\pgfsetstrokecolor{currentstroke}%
\pgfsetdash{}{0pt}%
\pgfsys@defobject{currentmarker}{\pgfqpoint{-0.041667in}{-0.041667in}}{\pgfqpoint{0.041667in}{0.041667in}}{%
\pgfpathmoveto{\pgfqpoint{0.000000in}{-0.041667in}}%
\pgfpathcurveto{\pgfqpoint{0.011050in}{-0.041667in}}{\pgfqpoint{0.021649in}{-0.037276in}}{\pgfqpoint{0.029463in}{-0.029463in}}%
\pgfpathcurveto{\pgfqpoint{0.037276in}{-0.021649in}}{\pgfqpoint{0.041667in}{-0.011050in}}{\pgfqpoint{0.041667in}{0.000000in}}%
\pgfpathcurveto{\pgfqpoint{0.041667in}{0.011050in}}{\pgfqpoint{0.037276in}{0.021649in}}{\pgfqpoint{0.029463in}{0.029463in}}%
\pgfpathcurveto{\pgfqpoint{0.021649in}{0.037276in}}{\pgfqpoint{0.011050in}{0.041667in}}{\pgfqpoint{0.000000in}{0.041667in}}%
\pgfpathcurveto{\pgfqpoint{-0.011050in}{0.041667in}}{\pgfqpoint{-0.021649in}{0.037276in}}{\pgfqpoint{-0.029463in}{0.029463in}}%
\pgfpathcurveto{\pgfqpoint{-0.037276in}{0.021649in}}{\pgfqpoint{-0.041667in}{0.011050in}}{\pgfqpoint{-0.041667in}{0.000000in}}%
\pgfpathcurveto{\pgfqpoint{-0.041667in}{-0.011050in}}{\pgfqpoint{-0.037276in}{-0.021649in}}{\pgfqpoint{-0.029463in}{-0.029463in}}%
\pgfpathcurveto{\pgfqpoint{-0.021649in}{-0.037276in}}{\pgfqpoint{-0.011050in}{-0.041667in}}{\pgfqpoint{0.000000in}{-0.041667in}}%
\pgfpathlineto{\pgfqpoint{0.000000in}{-0.041667in}}%
\pgfpathclose%
\pgfusepath{stroke,fill}%
}%
\begin{pgfscope}%
\pgfsys@transformshift{5.374164in}{1.123503in}%
\pgfsys@useobject{currentmarker}{}%
\end{pgfscope}%
\end{pgfscope}%
\begin{pgfscope}%
\definecolor{textcolor}{rgb}{0.000000,0.000000,0.000000}%
\pgfsetstrokecolor{textcolor}%
\pgfsetfillcolor{textcolor}%
\pgftext[x=5.774164in,y=1.045725in,left,base]{\color{textcolor}{\rmfamily\fontsize{16.000000}{19.200000}\selectfont\catcode`\^=\active\def^{\ifmmode\sp\else\^{}\fi}\catcode`\%=\active\def
\end{pgfscope}%
\begin{pgfscope}%
\pgfsetbuttcap%
\pgfsetroundjoin%
\definecolor{currentfill}{rgb}{0.000000,0.000000,0.000000}%
\pgfsetfillcolor{currentfill}%
\pgfsetlinewidth{0.803000pt}%
\definecolor{currentstroke}{rgb}{0.000000,0.000000,0.000000}%
\pgfsetstrokecolor{currentstroke}%
\pgfsetdash{}{0pt}%
\pgfsys@defobject{currentmarker}{\pgfqpoint{0.000000in}{0.000000in}}{\pgfqpoint{0.048611in}{0.000000in}}{%
\pgfpathmoveto{\pgfqpoint{0.000000in}{0.000000in}}%
\pgfpathlineto{\pgfqpoint{0.048611in}{0.000000in}}%
\pgfusepath{stroke,fill}%
}%
\begin{pgfscope}%
\pgfsys@transformshift{7.027052in}{1.035976in}%
\pgfsys@useobject{currentmarker}{}%
\end{pgfscope}%
\end{pgfscope}%
\begin{pgfscope}%
\definecolor{textcolor}{rgb}{0.000000,0.000000,0.000000}%
\pgfsetstrokecolor{textcolor}%
\pgfsetfillcolor{textcolor}%
\pgftext[x=7.124275in, y=0.936254in, left, base]{\color{textcolor}{\rmfamily\fontsize{20.000000}{24.000000}\selectfont\catcode`\^=\active\def^{\ifmmode\sp\else\^{}\fi}\catcode`\%=\active\def
\end{pgfscope}%
\begin{pgfscope}%
\pgfsetbuttcap%
\pgfsetroundjoin%
\definecolor{currentfill}{rgb}{0.000000,0.000000,0.000000}%
\pgfsetfillcolor{currentfill}%
\pgfsetlinewidth{0.803000pt}%
\definecolor{currentstroke}{rgb}{0.000000,0.000000,0.000000}%
\pgfsetstrokecolor{currentstroke}%
\pgfsetdash{}{0pt}%
\pgfsys@defobject{currentmarker}{\pgfqpoint{0.000000in}{0.000000in}}{\pgfqpoint{0.048611in}{0.000000in}}{%
\pgfpathmoveto{\pgfqpoint{0.000000in}{0.000000in}}%
\pgfpathlineto{\pgfqpoint{0.048611in}{0.000000in}}%
\pgfusepath{stroke,fill}%
}%
\begin{pgfscope}%
\pgfsys@transformshift{7.027052in}{1.789609in}%
\pgfsys@useobject{currentmarker}{}%
\end{pgfscope}%
\end{pgfscope}%
\begin{pgfscope}%
\definecolor{textcolor}{rgb}{0.000000,0.000000,0.000000}%
\pgfsetstrokecolor{textcolor}%
\pgfsetfillcolor{textcolor}%
\pgftext[x=7.124275in, y=1.689887in, left, base]{\color{textcolor}{\rmfamily\fontsize{20.000000}{24.000000}\selectfont\catcode`\^=\active\def^{\ifmmode\sp\else\^{}\fi}\catcode`\%=\active\def
\end{pgfscope}%
\begin{pgfscope}%
\pgfsetbuttcap%
\pgfsetroundjoin%
\definecolor{currentfill}{rgb}{0.000000,0.000000,0.000000}%
\pgfsetfillcolor{currentfill}%
\pgfsetlinewidth{0.803000pt}%
\definecolor{currentstroke}{rgb}{0.000000,0.000000,0.000000}%
\pgfsetstrokecolor{currentstroke}%
\pgfsetdash{}{0pt}%
\pgfsys@defobject{currentmarker}{\pgfqpoint{0.000000in}{0.000000in}}{\pgfqpoint{0.048611in}{0.000000in}}{%
\pgfpathmoveto{\pgfqpoint{0.000000in}{0.000000in}}%
\pgfpathlineto{\pgfqpoint{0.048611in}{0.000000in}}%
\pgfusepath{stroke,fill}%
}%
\begin{pgfscope}%
\pgfsys@transformshift{7.027052in}{2.543242in}%
\pgfsys@useobject{currentmarker}{}%
\end{pgfscope}%
\end{pgfscope}%
\begin{pgfscope}%
\definecolor{textcolor}{rgb}{0.000000,0.000000,0.000000}%
\pgfsetstrokecolor{textcolor}%
\pgfsetfillcolor{textcolor}%
\pgftext[x=7.124275in, y=2.443520in, left, base]{\color{textcolor}{\rmfamily\fontsize{20.000000}{24.000000}\selectfont\catcode`\^=\active\def^{\ifmmode\sp\else\^{}\fi}\catcode`\%=\active\def
\end{pgfscope}%
\begin{pgfscope}%
\pgfsetbuttcap%
\pgfsetroundjoin%
\definecolor{currentfill}{rgb}{0.000000,0.000000,0.000000}%
\pgfsetfillcolor{currentfill}%
\pgfsetlinewidth{0.803000pt}%
\definecolor{currentstroke}{rgb}{0.000000,0.000000,0.000000}%
\pgfsetstrokecolor{currentstroke}%
\pgfsetdash{}{0pt}%
\pgfsys@defobject{currentmarker}{\pgfqpoint{0.000000in}{0.000000in}}{\pgfqpoint{0.048611in}{0.000000in}}{%
\pgfpathmoveto{\pgfqpoint{0.000000in}{0.000000in}}%
\pgfpathlineto{\pgfqpoint{0.048611in}{0.000000in}}%
\pgfusepath{stroke,fill}%
}%
\begin{pgfscope}%
\pgfsys@transformshift{7.027052in}{3.296875in}%
\pgfsys@useobject{currentmarker}{}%
\end{pgfscope}%
\end{pgfscope}%
\begin{pgfscope}%
\definecolor{textcolor}{rgb}{0.000000,0.000000,0.000000}%
\pgfsetstrokecolor{textcolor}%
\pgfsetfillcolor{textcolor}%
\pgftext[x=7.124275in, y=3.197153in, left, base]{\color{textcolor}{\rmfamily\fontsize{20.000000}{24.000000}\selectfont\catcode`\^=\active\def^{\ifmmode\sp\else\^{}\fi}\catcode`\%=\active\def
\end{pgfscope}%
\begin{pgfscope}%
\pgfsetbuttcap%
\pgfsetroundjoin%
\definecolor{currentfill}{rgb}{0.000000,0.000000,0.000000}%
\pgfsetfillcolor{currentfill}%
\pgfsetlinewidth{0.803000pt}%
\definecolor{currentstroke}{rgb}{0.000000,0.000000,0.000000}%
\pgfsetstrokecolor{currentstroke}%
\pgfsetdash{}{0pt}%
\pgfsys@defobject{currentmarker}{\pgfqpoint{0.000000in}{0.000000in}}{\pgfqpoint{0.048611in}{0.000000in}}{%
\pgfpathmoveto{\pgfqpoint{0.000000in}{0.000000in}}%
\pgfpathlineto{\pgfqpoint{0.048611in}{0.000000in}}%
\pgfusepath{stroke,fill}%
}%
\begin{pgfscope}%
\pgfsys@transformshift{7.027052in}{4.050508in}%
\pgfsys@useobject{currentmarker}{}%
\end{pgfscope}%
\end{pgfscope}%
\begin{pgfscope}%
\definecolor{textcolor}{rgb}{0.000000,0.000000,0.000000}%
\pgfsetstrokecolor{textcolor}%
\pgfsetfillcolor{textcolor}%
\pgftext[x=7.124275in, y=3.950785in, left, base]{\color{textcolor}{\rmfamily\fontsize{20.000000}{24.000000}\selectfont\catcode`\^=\active\def^{\ifmmode\sp\else\^{}\fi}\catcode`\%=\active\def
\end{pgfscope}%
\begin{pgfscope}%
\pgfsetbuttcap%
\pgfsetroundjoin%
\definecolor{currentfill}{rgb}{0.000000,0.000000,0.000000}%
\pgfsetfillcolor{currentfill}%
\pgfsetlinewidth{0.803000pt}%
\definecolor{currentstroke}{rgb}{0.000000,0.000000,0.000000}%
\pgfsetstrokecolor{currentstroke}%
\pgfsetdash{}{0pt}%
\pgfsys@defobject{currentmarker}{\pgfqpoint{0.000000in}{0.000000in}}{\pgfqpoint{0.048611in}{0.000000in}}{%
\pgfpathmoveto{\pgfqpoint{0.000000in}{0.000000in}}%
\pgfpathlineto{\pgfqpoint{0.048611in}{0.000000in}}%
\pgfusepath{stroke,fill}%
}%
\begin{pgfscope}%
\pgfsys@transformshift{7.027052in}{4.804141in}%
\pgfsys@useobject{currentmarker}{}%
\end{pgfscope}%
\end{pgfscope}%
\begin{pgfscope}%
\definecolor{textcolor}{rgb}{0.000000,0.000000,0.000000}%
\pgfsetstrokecolor{textcolor}%
\pgfsetfillcolor{textcolor}%
\pgftext[x=7.124275in, y=4.704418in, left, base]{\color{textcolor}{\rmfamily\fontsize{20.000000}{24.000000}\selectfont\catcode`\^=\active\def^{\ifmmode\sp\else\^{}\fi}\catcode`\%=\active\def
\end{pgfscope}%
\begin{pgfscope}%
\pgfsetbuttcap%
\pgfsetroundjoin%
\definecolor{currentfill}{rgb}{0.000000,0.000000,0.000000}%
\pgfsetfillcolor{currentfill}%
\pgfsetlinewidth{0.803000pt}%
\definecolor{currentstroke}{rgb}{0.000000,0.000000,0.000000}%
\pgfsetstrokecolor{currentstroke}%
\pgfsetdash{}{0pt}%
\pgfsys@defobject{currentmarker}{\pgfqpoint{0.000000in}{0.000000in}}{\pgfqpoint{0.048611in}{0.000000in}}{%
\pgfpathmoveto{\pgfqpoint{0.000000in}{0.000000in}}%
\pgfpathlineto{\pgfqpoint{0.048611in}{0.000000in}}%
\pgfusepath{stroke,fill}%
}%
\begin{pgfscope}%
\pgfsys@transformshift{7.027052in}{5.557774in}%
\pgfsys@useobject{currentmarker}{}%
\end{pgfscope}%
\end{pgfscope}%
\begin{pgfscope}%
\definecolor{textcolor}{rgb}{0.000000,0.000000,0.000000}%
\pgfsetstrokecolor{textcolor}%
\pgfsetfillcolor{textcolor}%
\pgftext[x=7.124275in, y=5.458051in, left, base]{\color{textcolor}{\rmfamily\fontsize{20.000000}{24.000000}\selectfont\catcode`\^=\active\def^{\ifmmode\sp\else\^{}\fi}\catcode`\%=\active\def
\end{pgfscope}%
\begin{pgfscope}%
\definecolor{textcolor}{rgb}{0.000000,0.000000,0.000000}%
\pgfsetstrokecolor{textcolor}%
\pgfsetfillcolor{textcolor}%
\pgftext[x=7.633997in,y=3.211528in,,top,rotate=90.000000]{\color{textcolor}{\rmfamily\fontsize{24.000000}{28.800000}\selectfont\catcode`\^=\active\def^{\ifmmode\sp\else\^{}\fi}\catcode`\%=\active\def
\end{pgfscope}%
\begin{pgfscope}%
\pgfpathrectangle{\pgfqpoint{0.893781in}{0.789058in}}{\pgfqpoint{6.133272in}{4.844938in}}%
\pgfusepath{clip}%
\pgfsetrectcap%
\pgfsetroundjoin%
\pgfsetlinewidth{1.505625pt}%
\definecolor{currentstroke}{rgb}{1.000000,0.000000,0.000000}%
\pgfsetstrokecolor{currentstroke}%
\pgfsetdash{}{0pt}%
\pgfpathmoveto{\pgfqpoint{1.172566in}{1.624918in}}%
\pgfpathlineto{\pgfqpoint{2.566491in}{1.009283in}}%
\pgfpathlineto{\pgfqpoint{3.960417in}{1.328301in}}%
\pgfpathlineto{\pgfqpoint{5.354342in}{2.261217in}}%
\pgfpathlineto{\pgfqpoint{6.748267in}{5.413772in}}%
\pgfusepath{stroke}%
\end{pgfscope}%
\begin{pgfscope}%
\pgfpathrectangle{\pgfqpoint{0.893781in}{0.789058in}}{\pgfqpoint{6.133272in}{4.844938in}}%
\pgfusepath{clip}%
\pgfsetbuttcap%
\pgfsetroundjoin%
\definecolor{currentfill}{rgb}{1.000000,0.000000,0.000000}%
\pgfsetfillcolor{currentfill}%
\pgfsetlinewidth{1.003750pt}%
\definecolor{currentstroke}{rgb}{1.000000,0.000000,0.000000}%
\pgfsetstrokecolor{currentstroke}%
\pgfsetdash{}{0pt}%
\pgfsys@defobject{currentmarker}{\pgfqpoint{-0.041667in}{-0.041667in}}{\pgfqpoint{0.041667in}{0.041667in}}{%
\pgfpathmoveto{\pgfqpoint{0.000000in}{-0.041667in}}%
\pgfpathcurveto{\pgfqpoint{0.011050in}{-0.041667in}}{\pgfqpoint{0.021649in}{-0.037276in}}{\pgfqpoint{0.029463in}{-0.029463in}}%
\pgfpathcurveto{\pgfqpoint{0.037276in}{-0.021649in}}{\pgfqpoint{0.041667in}{-0.011050in}}{\pgfqpoint{0.041667in}{0.000000in}}%
\pgfpathcurveto{\pgfqpoint{0.041667in}{0.011050in}}{\pgfqpoint{0.037276in}{0.021649in}}{\pgfqpoint{0.029463in}{0.029463in}}%
\pgfpathcurveto{\pgfqpoint{0.021649in}{0.037276in}}{\pgfqpoint{0.011050in}{0.041667in}}{\pgfqpoint{0.000000in}{0.041667in}}%
\pgfpathcurveto{\pgfqpoint{-0.011050in}{0.041667in}}{\pgfqpoint{-0.021649in}{0.037276in}}{\pgfqpoint{-0.029463in}{0.029463in}}%
\pgfpathcurveto{\pgfqpoint{-0.037276in}{0.021649in}}{\pgfqpoint{-0.041667in}{0.011050in}}{\pgfqpoint{-0.041667in}{0.000000in}}%
\pgfpathcurveto{\pgfqpoint{-0.041667in}{-0.011050in}}{\pgfqpoint{-0.037276in}{-0.021649in}}{\pgfqpoint{-0.029463in}{-0.029463in}}%
\pgfpathcurveto{\pgfqpoint{-0.021649in}{-0.037276in}}{\pgfqpoint{-0.011050in}{-0.041667in}}{\pgfqpoint{0.000000in}{-0.041667in}}%
\pgfpathlineto{\pgfqpoint{0.000000in}{-0.041667in}}%
\pgfpathclose%
\pgfusepath{stroke,fill}%
}%
\begin{pgfscope}%
\pgfsys@transformshift{1.172566in}{1.624918in}%
\pgfsys@useobject{currentmarker}{}%
\end{pgfscope}%
\begin{pgfscope}%
\pgfsys@transformshift{2.566491in}{1.009283in}%
\pgfsys@useobject{currentmarker}{}%
\end{pgfscope}%
\begin{pgfscope}%
\pgfsys@transformshift{3.960417in}{1.328301in}%
\pgfsys@useobject{currentmarker}{}%
\end{pgfscope}%
\begin{pgfscope}%
\pgfsys@transformshift{5.354342in}{2.261217in}%
\pgfsys@useobject{currentmarker}{}%
\end{pgfscope}%
\begin{pgfscope}%
\pgfsys@transformshift{6.748267in}{5.413772in}%
\pgfsys@useobject{currentmarker}{}%
\end{pgfscope}%
\end{pgfscope}%
\begin{pgfscope}%
\pgfsetrectcap%
\pgfsetmiterjoin%
\pgfsetlinewidth{0.803000pt}%
\definecolor{currentstroke}{rgb}{0.000000,0.000000,0.000000}%
\pgfsetstrokecolor{currentstroke}%
\pgfsetdash{}{0pt}%
\pgfpathmoveto{\pgfqpoint{0.893781in}{0.789058in}}%
\pgfpathlineto{\pgfqpoint{0.893781in}{5.633997in}}%
\pgfusepath{stroke}%
\end{pgfscope}%
\begin{pgfscope}%
\pgfsetrectcap%
\pgfsetmiterjoin%
\pgfsetlinewidth{0.803000pt}%
\definecolor{currentstroke}{rgb}{0.000000,0.000000,0.000000}%
\pgfsetstrokecolor{currentstroke}%
\pgfsetdash{}{0pt}%
\pgfpathmoveto{\pgfqpoint{7.027052in}{0.789058in}}%
\pgfpathlineto{\pgfqpoint{7.027052in}{5.633997in}}%
\pgfusepath{stroke}%
\end{pgfscope}%
\begin{pgfscope}%
\pgfsetrectcap%
\pgfsetmiterjoin%
\pgfsetlinewidth{0.803000pt}%
\definecolor{currentstroke}{rgb}{0.000000,0.000000,0.000000}%
\pgfsetstrokecolor{currentstroke}%
\pgfsetdash{}{0pt}%
\pgfpathmoveto{\pgfqpoint{0.893781in}{0.789058in}}%
\pgfpathlineto{\pgfqpoint{7.027053in}{0.789058in}}%
\pgfusepath{stroke}%
\end{pgfscope}%
\begin{pgfscope}%
\pgfsetrectcap%
\pgfsetmiterjoin%
\pgfsetlinewidth{0.803000pt}%
\definecolor{currentstroke}{rgb}{0.000000,0.000000,0.000000}%
\pgfsetstrokecolor{currentstroke}%
\pgfsetdash{}{0pt}%
\pgfpathmoveto{\pgfqpoint{0.893781in}{5.633997in}}%
\pgfpathlineto{\pgfqpoint{7.027053in}{5.633997in}}%
\pgfusepath{stroke}%
\end{pgfscope}%
\end{pgfpicture}%
\makeatother%
\endgroup%

%% file: Sections/6_conclusion.tex
It is concluded that it is possible to perform \acrlong{MIA}s on \acrlong{LDM}s fine-tuned on face images. The attack proposed is restricted to the task of inferring membership on a dataset level, and is not successful in the task of inferring which specific images of a dataset are used for fine-tuning a target model.\\
The circumstances or prerequisites for the cases where the MIA is successful are investigated and multiple factors are considered in relation to its performance. It is found that using a generated auxiliary dataset leads to a significantly better performance for the MIA than using real images. Diluting the members of interest by mixing two datasets in the fine-tuning of the target model did not lead to significantly worse performance of the MIA. It should however be noted that this was only tested by reducing the share of target members to 1:2. \\
It is found that by introducing visible watermarks to the target dataset, our MIA sees a significant boost in performance. Using hidden watermarks was not found to have a positive impact on the performance of the MIA. No significant effect was found when investigating the influence of the relationship between the labels used for fine-tuning the target model and the prompt used for inference, i.e. whether they match or not. Upon investigating the importance of the guidance scale used by the target model, it is found to have a significant influence on the performance of our MIA, with best performance at $s\sim8$.\\ 
Overall the proposed MIA is a realistic and feasible attack in a real-life application. However, it is computationally expensive to fine-tune a generative "shadow model" for the task of producing an auxiliary dataset related to the domain of interest as well as training the Resnet-18 attack model. The nature of the tests performed restricts our conclusion to the case of LDMs fine-tuned on face images. The smallest amount of fine-tuning that was still found to be effective was 50 epochs on the member images (however it could be lower - as it was not tested). The only LDM used for testing was Stable-Diffusion-v1.5 \cite{rombach2022highresolution}, which limits the generalizability of the conclusions drawn. The approach using a Resnet-18 as an attack model is found to be generally stable on several different hyperparameters in the target LDM. In conclusion, the method for \acrlong{MIA} shown in this paper is realistic and could be used as a tool to infer if one's face images have been used to fine-tune a \acrlong{LDM} in a black-box setup.